\pdfoutput=1

\documentclass[11pt]{article}

\usepackage[final]{ACL2023}

\usepackage{times}
\usepackage{latexsym}

\usepackage{graphicx}
\usepackage{subfigure}
\usepackage{enumitem}
\usepackage{xcolor}

\usepackage{graphicx}
\usepackage{caption}
\usepackage{subcaption}
\usepackage{amsmath}
\usepackage{amssymb}
\usepackage{CJKutf8}
\usepackage[linesnumbered,ruled,longend]{algorithm2e}
\usepackage{float}
\usepackage{tabularx}  
\usepackage{array}     
\usepackage{adjustbox} 
\usepackage{listings}
\lstdefinelanguage{json}{
    basicstyle=\footnotesize\ttfamily, 
    numbers=left,
    numberstyle=\tiny, 
    stepnumber=1,
    numbersep=5pt,
    showstringspaces=false,
    breaklines=true,
    frame=lines,
    backgroundcolor=\color{gray!10},
    literate=
     *{0}{{{\color{blue!60!black}0}}}{1}
      {1}{{{\color{blue!60!black}1}}}{1}
      {2}{{{\color{blue!60!black}2}}}{1}
      {3}{{{\color{blue!60!black}3}}}{1}
      {4}{{{\color{blue!60!black}4}}}{1}
      {5}{{{\color{blue!60!black}5}}}{1}
      {6}{{{\color{blue!60!black}6}}}{1}
      {7}{{{\color{blue!60!black}7}}}{1}
      {8}{{{\color{blue!60!black}8}}}{1}
      {9}{{{\color{blue!60!black}9}}}{1}
      {:}{{{\color{black}{:}}}}{1}
      {,}{{{\color{black}{,}}}}{1}
      {\{}{{{\color{black}{\{}}}}{1}
      {\}}{{{\color{black}{\}}}}}{1}
      {[}{{{\color{black}{[}}}}{1}
      {]}{{{\color{black}{]}}}}{1},
}

\usepackage[T1]{fontenc}

\usepackage[utf8]{inputenc}

\usepackage{microtype}

\usepackage{inconsolata}
\usepackage{booktabs}
\usepackage{multirow}
\usepackage{arydshln}

 \title{ An Efficient Task-Oriented Dialogue Policy: Evolutionary Reinforcement Learning Injected by Elite Individuals}

\author{
Yangyang Zhao$^{1}$ \quad Ben Niu$^{1}$ \quad Libo Qin$^{2*}$ \quad Shihan Wang$^{3}$ \\
$^{1}$School of  Computer Science and Technology, Changsha University of Science and Technology, China \\
$^{2}$School of Computer Science and Engineering, Central South University, China\\
$^{3}$School of  Information and Computing Sciences, Utrecht University, Netherlands \\
\texttt{yyz@csust.edu.cn; niuniuniu@stu.csust.edu.cn; lbqin@csu.edu.cn; s.wang2@uu.nl}
}

\begin{document}
\begin{CJK}{UTF8}{gbsn}
\maketitle

\begingroup
\renewcommand\thefootnote{*}  
\footnotetext{Corresponding author}
\endgroup
\begin{abstract}
Deep Reinforcement Learning (DRL) is widely used in task-oriented dialogue systems to optimize dialogue policy, but it struggles to balance exploration and exploitation due to the high dimensionality of state and action spaces. This challenge often results in local optima or poor convergence. Evolutionary Algorithms (EAs) have been proven to effectively explore the solution space of neural networks by maintaining population diversity. Inspired by this, we innovatively combine the global search capabilities of EA with the local optimization of DRL to achieve a balance between exploration and exploitation. Nevertheless, the inherent flexibility of natural language in dialogue tasks complicates this direct integration, leading to prolonged evolutionary times. Thus, we further propose an elite individual injection mechanism to enhance EA's search efficiency by adaptively introducing best-performing individuals into the population. Experiments across four datasets show that our approach significantly improves the balance between exploration and exploitation,  boosting performance. Moreover, the effectiveness of the EII mechanism in reducing exploration time has been demonstrated, achieving an efficient integration of EA and DRL on task-oriented dialogue policy tasks.

\end{abstract}

\section{Introduction}
Task-oriented dialogue (TOD) systems aim to understand user intent \citep{qin2021survey}, generate responses \citep{qin2023end}, and steer conversations toward goals, with the dialogue policy (DP) \citep{kwan2019survey} playing a crucial role in determining and steering conversations. Deep Reinforcement Learning (DRL) has proven effective in optimizing DPs due to its strengths in sequential decision-making \citep{kwan2023survey}. While Large Language Models \citep{chung2023instructtods, hu2024dialight, kumar2024large} are increasingly used in TOD systems for their language capabilities, they are rarely applied in DPs because of their limited decision-making ability and the high resource demands for prompt design and fine-tuning \citep{yao2024tree, Yang2024Harnessing}. Thus, optimizing DPs through DRL remains the primary approach \citep{du2024rewarding}.  However, DRL struggles to balance exploration and exploitation. Insufficient exploration can lead to local optima, while excessive exploration results in inefficient training\footnote{We have shown that the above observation also applies to the dialogue task, with experimental results and analysis provided in \autoref{app:explor}.} \citep{henderson2018deep}.

\begin{figure*}[h]
	\centering
	\includegraphics[width=0.85\linewidth]{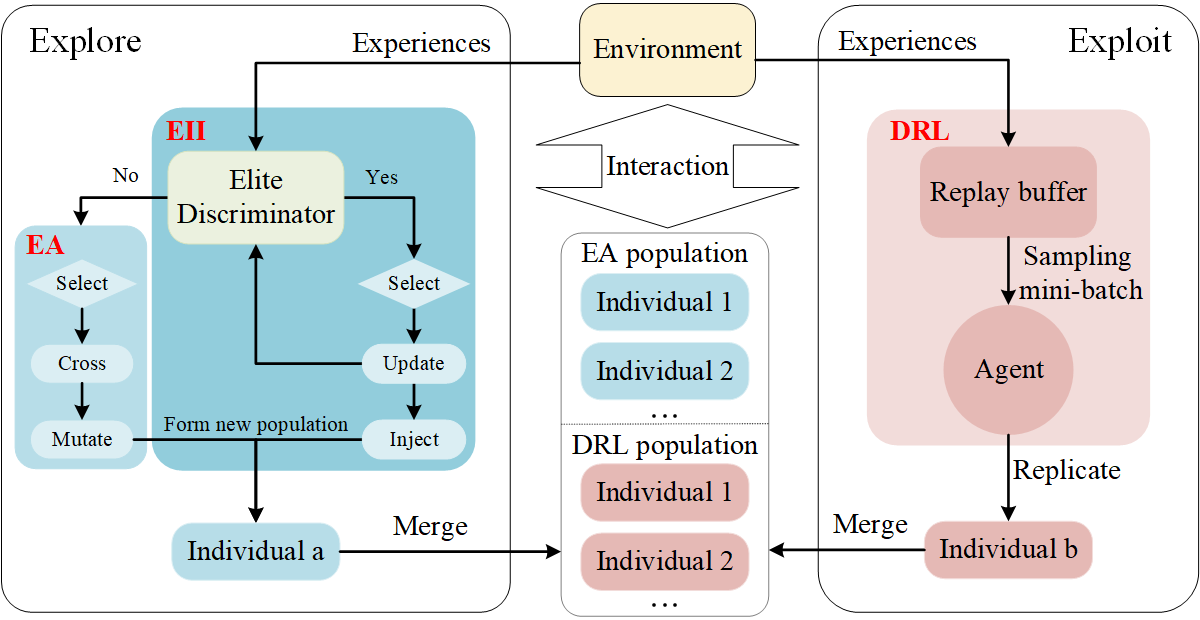}
	\caption{Framework of the EIERL method, comprising two modules: exploration and exploitation. The exploration module aims to form an EA population (multiple individuals a) to explore diverse experiences, while the exploitation module aims to form a DRL population (multiple individuals b) to use existing experiences for training.}
	\label{fig:figure1}
\end{figure*}

Research on balancing exploration and exploitation in DRL-based DP can be categorized into two main approaches: direct and indirect enhancement of exploration efficiency. Direct methods focus on designing effective exploration strategies that, while powerful, are often limited to specific domains and are time-consuming \citep{LadoszWKO22}. Indirect methods involve leveraging expert knowledge to guide exploration \citep{wang2020learning,xu2021improving} and developing high-quality user simulators to generate diverse experiences \citep{peng2018deep}. However, these approaches add costs for constructing expert knowledge and user simulators. 
Evolutionary Algorithms (EAs) are theoretically well-suited for balancing exploration and exploitation due to their population-based diversity exploration advantages \citep{CrepinsekLM13}. 

However, EAs struggle with exploitation because they lack gradient guidance, while gradient descent is a key strength of sample-efficient DRL approaches \citep{abs-2401-11963}. A natural solution is to combine the complementary strengths of EAs and DRL in task-oriented dialogue policy tasks, referred to as ERL-based DP, to achieve an effective balance between exploration and exploitation \citep{khadka2018evolution, Suri22}. However, this integration is not straightforward. The inherent flexibility of natural language in TOD tasks results in a vast search space, requiring significant time for EAs to evolve effective strategies(as demonstrated in \autoref{sec:result2}).

To address the above challenges, we propose an Elite Individual Injection (EII) mechanism, integrating it into an ERL-based DP algorithm to form a novel hybrid algorithm called EIERL. As shown in \autoref{fig:figure1}, EIERL comprises two components: Exploration (left) and Exploitation (right). In the \textbf{Exploration} phase, the EA population generates diverse strategies through selection, crossover, and mutation to explore various experiences, while the EII mechanism periodically injects best-performing individuals into the EA population to accelerate evolution. The injection timing is adaptively controlled by an elite discriminator, which adaptively selects optimal injection moments and updates based on the training stage. In the \textbf{Exploitation} phase, the DRL agent optimizes its performance using exsiting experiences and replicates optimized agents to form a DRL population. Both the formed EA and DRL populations then interact with the environment to generate diverse exploration experiences and exploitation experiences. The strengths and generalizability of EIERL in effectively trade-off exploration and exploitation are demonstrated on four public dialogue datasets. Moreover, the EII mechanism has proven effective in accelerating the EA's exploration process, enabling a seamless integration of DRL and EA for task-oriented dialogue policy tasks. To our knowledge, this is the first study to integrate EAs and DRL into task-oriented dialogue policy tasks.

In summary, our contributions are threefold:

\begin{itemize}
\vspace{-5pt}
\item We propose a hybrid algorithm that leverages the complementary strengths of EAs and DRL to effectively balance exploration and exploitation in dialogue policy learning. 
\vspace{-5pt}
\item We introduce an elite individual injection mechanism to solve the prolonged evolution challenges faced by traditional ERL directly migrating to dialogue task, enabling an effective combination of EA and DRL.
\vspace{-5pt}
\item  We validate the effectiveness and generalizability of EIERL in multiple dialogue datasets.
\end{itemize}

\vspace{-5pt}
\section{Related Work}

\subsection{The Exploration-Exploitation Trade-off}
Methods for balancing exploration and exploitation in DRL-based task-oriented dialogue policies are generally categorized into direct and indirect enhancements of exploration efficiency. 

Direct methods involve designing tailored exploration strategies, which are often labor-intensive \citep{LadoszWKO22}. For instance, \citet{tegho2017uncertainty} and \citet{lipton2016efficient} proposed a Bayes-by-Backprop approach based on Deep Q-Networks (DQN), which customizes exploration strategies by utilizing Thompson sampling to estimate uncertainty from Monte Carlo samples. However, the effectiveness of this method heavily relies on the choice of prior distributions, which may limit its applicability across different scenarios.

Indirect methods utilize expert knowledge to guide exploration \citep{xu2021improving,wang2020learning} or develop high-quality user simulators to generate diverse experiences \citep{peng2018deep}. However, expert-driven approaches depend heavily on data quality, where poor data can hinder exploration or lead to ineffective reward structures \citep{kang2018policy}. For example, \citet{nishimoto2019dialogue} combined a softmax strategy with transfer learning, using the source task as expert knowledge, and employed a Boltzmann distribution to assign probabilities to action-state pairs, thereby improving exploration effectiveness. Nevertheless, the differences between the source and target tasks in transfer learning restrict the model's adaptability. For user simulator-based methods, although \citet{peng2018deep} demonstrated that user simulators can enhance efficiency, they require substantial expertise and annotated data, and their quality can be difficult to assess due to challenges in replicating real user behavior \citep{kwan2023survey}.


\vspace{-4pt}
\subsection{ERL: Integration of DRL and EA}
\vspace{-3pt}

ERL combines the population diversity of EA with the gradient information from DRL, showing promise in addressing exploration-exploitation challenges \citep{khadka2018evolution}. This integration enhances sample efficiency in the game domain, but the introduction of multiple hyperparameters complicates tuning. To address this, \citet{dong2024adaptive} proposed Adaptive Evolutionary Reinforcement Learning (AERL), which dynamically adjusts the ratio of RL to EA populations based on performance, improving training efficiency and robustness. \citet{jianye2022erl} further introduced ERL-Re2, facilitating knowledge sharing between EA and RL agents by adjusting state representations while keeping policy representations independent. Traditional ERL methods often struggle with policy-based RL due to inefficient handling of value functions and sample usage. To improve sample efficiency, \citet{li2023value} suggested using negative temporal difference as a fitness metric, aligning with value iteration principles. 

\begin{figure*}[htbp]
\centering

\begin{minipage}[b]{4.8cm}
    \centering
    \includegraphics[width=5cm]{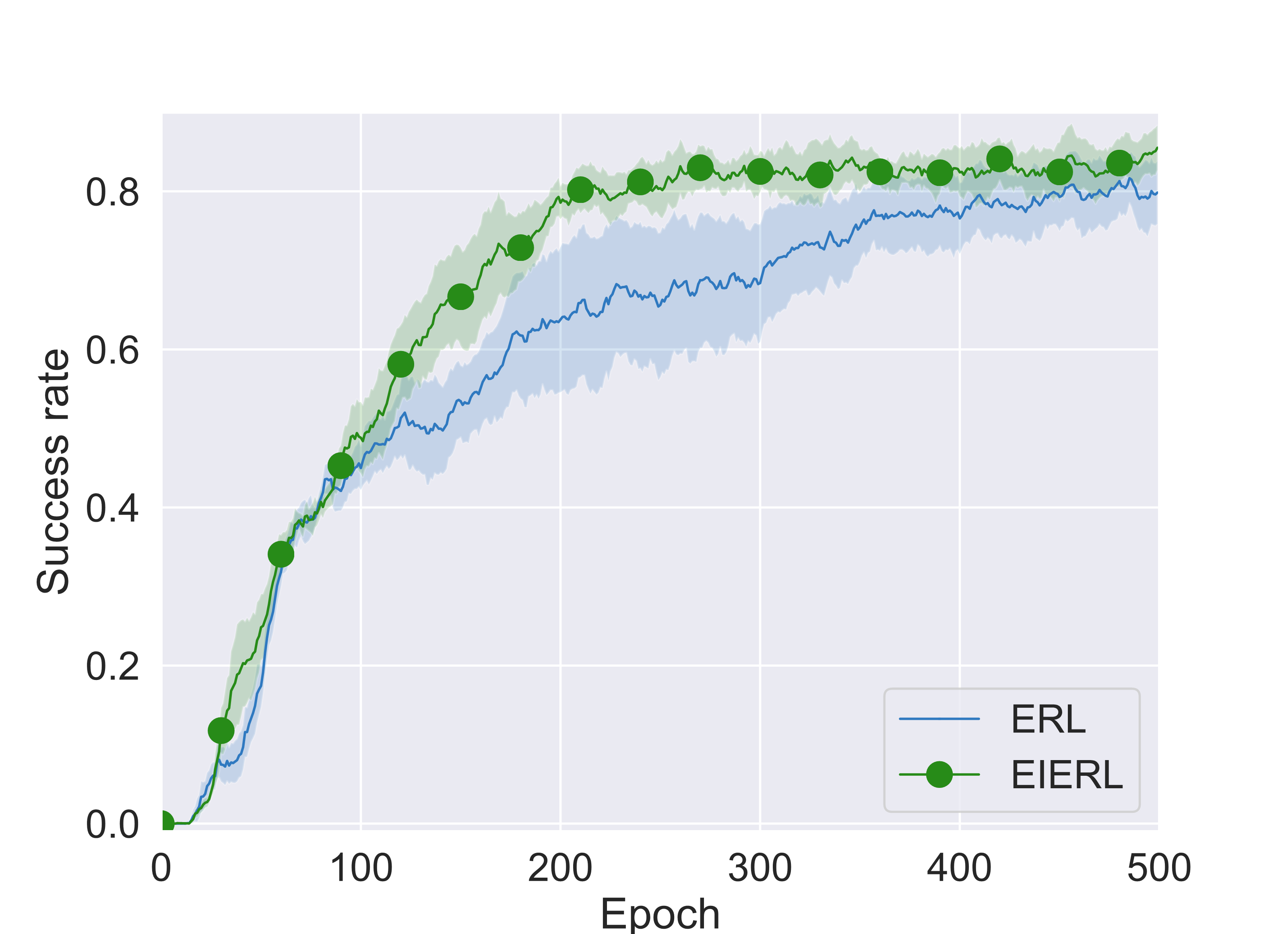}
    \caption*{(a) Movie}
\end{minipage}
\hspace{0.5cm}
\begin{minipage}[b]{4.8cm}
    \centering
    \includegraphics[width=5cm]{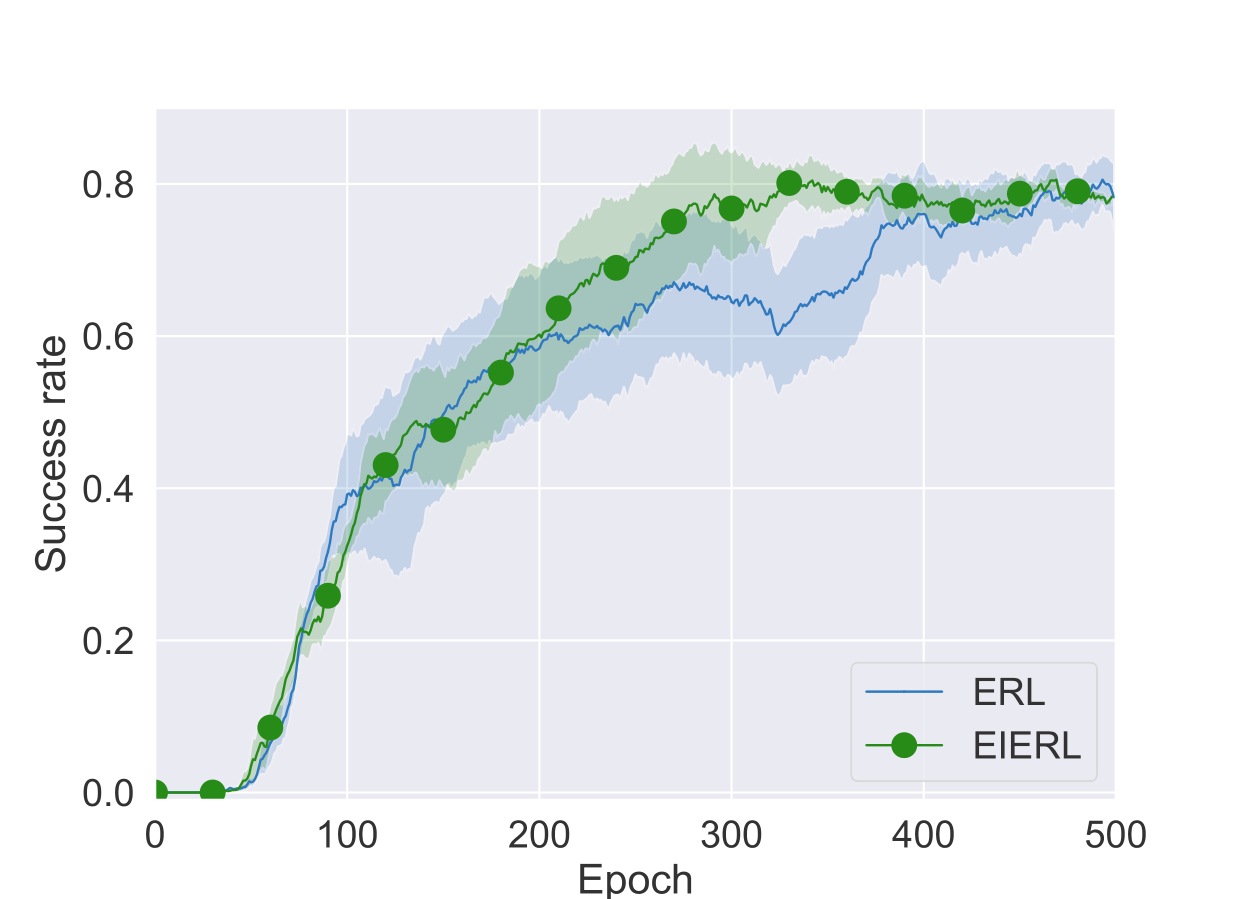}
    \caption*{(b) Rest}
\end{minipage}
\hspace{0.5cm}
\begin{minipage}[b]{4.8cm}
    \centering
    \includegraphics[width=5cm]{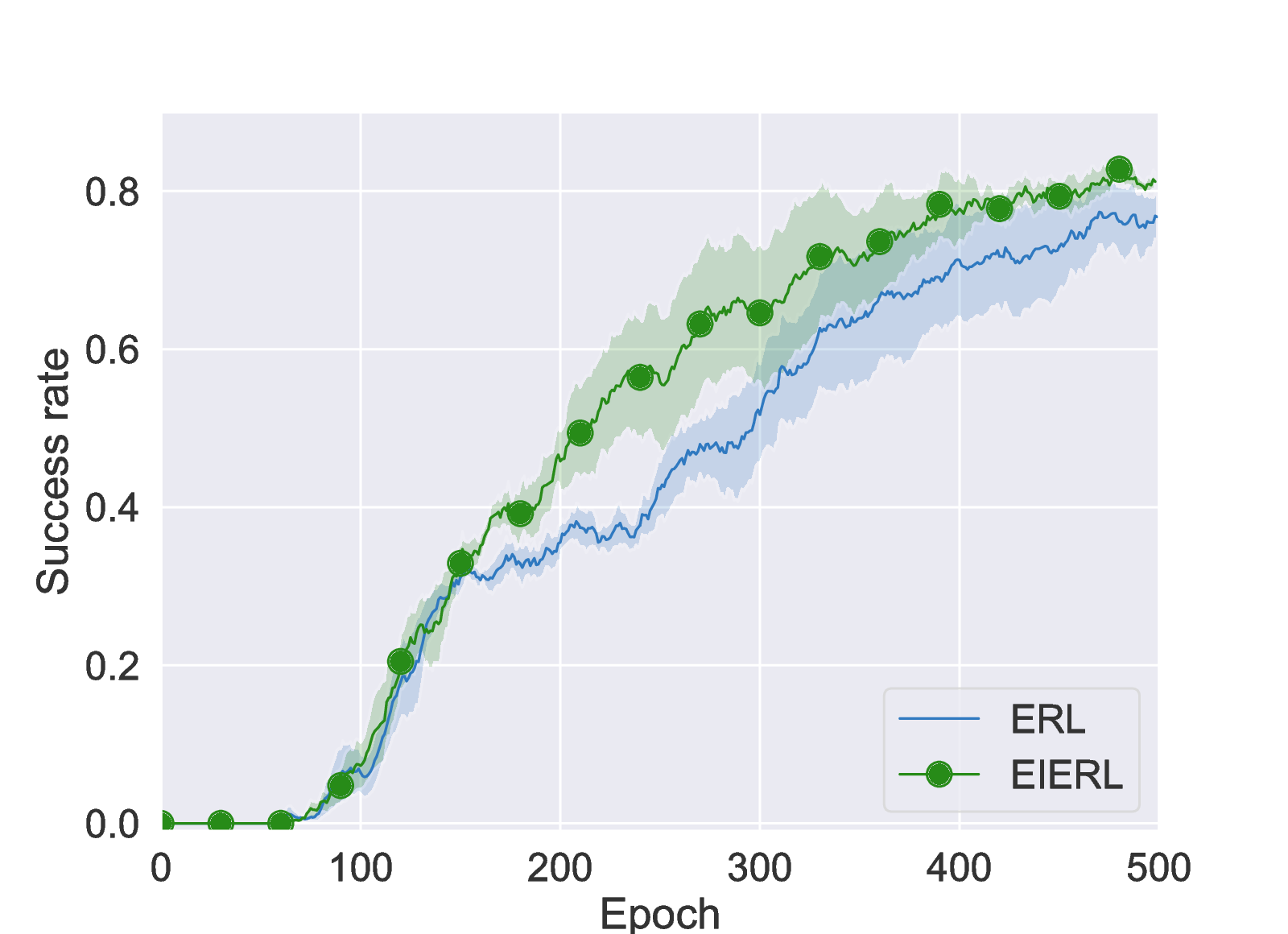}
    \caption*{(c) Taxi}
\end{minipage}

\caption{The effect of the EII mechanism on performance in Movie, Restaurant, and Taxi domains.}
\label{fig:figure4}
\end{figure*}

Despite these advancements, most ERL research has focused on the game tasks, with limited exploration of the dialogue tasks. Dialogue tasks involve flexible natural language, diverse user behaviors, and varying dialogue environments, resulting in a significantly larger search space compared to game tasks. Consequently, the EA module in ERL requires a considerable amount of time to evolve effective policies. Moreover, many explorations in this large search space are ineffective, hampering ERL's ability to identify reliable evolutionary directions and thus reducing learning efficiency. Our experimental results, as shown in \autoref{fig:figure4} and \hyperref[fig:figure8(b)]{Figure 6(b)}, corroborate this point. Inspired by this gap, we propose a novel approach that integrates EA with DRL-based dialogue policy methods while tackling the prolonged evolution times of traditional EA methods in dialogue tasks, enabling an efficient integration between EA and DRL.

\vspace{-4pt}
\section{EIERL}
\vspace{-3pt}
As illustrated in \autoref{fig:figure1}, EIERL consists of two modules: (1) the Exploitation module, implemented using a DRL algorithm, which trains the agents using existing experiences and replicates them to form a DRL population\footnote{In this paper, we utilize the commonly used DQN algorithm in DP tasks for experiments. However, the conceptual framework of EIERL applies to other offline RL-based DP methods.}; (2) the Exploration module, implemented by two submodules: EA and EII mechanism. The EA submodule performs selection, crossover, and mutation operations on individuals within both the EA and DRL populations to create the EA population, facilitating an in-depth exploration of the state space. Meanwhile, the EII submodule adaptively determines the optimal timing for injecting the best-performing individuals into the EA population, thereby accelerating the evolutionary process. Finally, EIERL integrates the DRL and EA populations to interact with the environment, generating experiences for both exploitation and exploration. The full procedure of the EIERL is described in \autoref{alg:eierl}.

\subsection{Exploitation Module}
The exploitation module aims to create a DRL population by replicating agents optimized using existing experiences stored in an experience buffer. Specifically, the agents sample a mini-batch of experiences from the experience replay buffer $\mathcal{D}$ and update the parameters $\theta_Q$ of the value function $Q(\cdot)$ by minimizing the mean-squared loss function:

\vspace{-6mm}
\begin{equation}  
\label{gongshi1}
\mathcal{L}(\theta_Q) = \mathbb{E}_{(s,a,r,s')\sim \mathcal{D}}[(y_i - Q_{\theta_Q}(s,a))^2]
\end{equation}
where $y_i$ is specified as follows:

\vspace{-2mm}
\begin{equation}  
\label{gongshi2}
y_i = r+\gamma \max_{a'}Q'_{\theta_{Q'}}(s',a')
\end{equation}

where the experiences stored in $\mathcal{D}$ are in the form of (state $s$,action $a$,reward $r$, next$\_${state} $s'$). $\gamma$ denotes the discount factor and satisfies $0 \le \gamma \le 1$. The target network $Q'_{\theta_{Q'}}$ is refreshed at scheduled intervals, whereas the online network $Q_{\theta_Q}$ is trained by backpropagation using mini batch gradient descent.

The optimized agents are then replicated to form DRL individuals (individual b in \autoref{fig:figure1}), which together constitute the DRL population.

\subsection{Exploration Module}
The exploration module focuses on maintaining a diverse EA population through EA to enhance exploration efficiency by generating varied experiences. In dialogue tasks with large state spaces, traditional EA often begins with a low-fitness population, leading to inefficient exploration of suboptimal areas and high computational costs. To improve this, we introduced the EII mechanism, which uses an elite discriminator to determine the suitable timing for injecting best-performing individuals into the EA population, guiding the search toward more optimal directions and enhancing the exploration process.

\subsubsection{EAs}

The procedure of the EAs is detailed in \autoref{alg:evolution}. It starts with tournament selection \citep{goldberg1991comparative} to prioritize individuals from regions with higher potential returns \citep{khadka2018evolution}. Next, gene crossover occurs within the population, followed by mutations applied based on specific probabilities, systematically altering individuals' genetic structures to improve evolution. These mutations adjust neural network weights using a normal distribution.
\subsubsection{EII mechanism}

\begin{algorithm}[h]

\DontPrintSemicolon
\caption{Function Evolution}
\label{alg:evolution}

\SetKwFunction{Evolution} {Evolution}
\SetKwProg{Fn}{Function}{:}{}
\Fn{\Evolution{$pop_{evo}$,all\_fitness}}{

    rank population by $all\_fitness$\;
    select best $e = \text{int}(\psi \cdot m)$ individuals as elite\;
    select $(m-e)$ individuals from $pop_{evo}$ using tournament selection to form $S$\;

    \While{\(|S| < (m - e)\)}{
    crossover between a random $\pi \in e$ and $\pi \in S$, append to $S$\;
}

 \For{$\pi \in S$}{
    \If{$r() < \text{mut}_{\text{prob}}$}{
        \For{$\mathcal{M} \in \theta_\pi$}{
            perform $\text{mut}_{\text{frac}} \cdot |\mathcal{M}|$ mutations\;
            sample indices $i, j$ from $\mathcal{M}$\;
            
            \If{$r() < \text{supermut}_{\text{prob}}$}{
                $\mathcal{M}[i, j] \cdot= \mathcal{N}(0, 100 \cdot \text{mut}_{\text{strength}})$\;
            }
            \ElseIf{$r() < \text{reset}_{\text{prob}}$}{
                $\mathcal{M}[i, j] = \mathcal{N}(0, 1)$\;
            }
            \Else{
                $\mathcal{M}[i, j] \cdot= \mathcal{N}(0, \text{mut}_{\text{strength}})$\;
            }
        }
    }
}

    \Return{ $pop_{evo}$}\
}
\end{algorithm}
The EII mechanism is outlined in lines 9-21 of \autoref{alg:eierl2}, where the elite discriminator adaptively determines the timing for elite individual injection. The discriminator evaluates the fitness of each individual by receiving feedback from their interactions with the environment. Specifically, fitness is measured by the cumulative reward obtained from these interactions, with the calculation process detailed in \autoref{alg:evaluate}. When an individual's fitness exceeds the threshold $f_{max}$ (initialized as $-\infty$), it signifies the emergence of an elite individual during the current iteration, and the elite injection process is triggered. The individual achieving the maximum fitness value, $f_{max}^{'}$, is then selected as the elite individual, $\pi_{max}$. Since the fitness threshold reflects the agent's learning progress, the discriminator updates $f_{max}$ to $f_{max}^{'}$, allowing it to adaptively raise the standard as learning progresses. Finally, the elite individual $\pi_{max}$ is injected into the EA population (individual a in \autoref{fig:figure1}). If no elite individual is detected, only the EA process is executed.

\begin{algorithm}
\caption{Function Evaluate}
\label{alg:evaluate}
\DontPrintSemicolon
\SetKwFunction{Evaluate} {Evaluate}
\SetKwProg{Fn}{Function}{:}{}
\Fn{\Evaluate{$\pi$,$\mathcal{D}$}}{
    $fitness = 0$\
    
     \For{$i = 1$ to $\xi$}{
        initialize environment and state $s_0$

        \While{dialogue is not done}{
           \If{rand() < $\epsilon$}{
    $a_t$ = random  action
}
\Else{
   $a_t = argmax_{a'} Q(s_t, a'; \theta_{\pi})$
}
           execute $a_t$, observe $r_t$ and $s_{t+1}$ \\
append $(s_t, a_t, r_t, s_{t+1})$ to $\mathcal{D}$

            $fitness = fitness + r_t$ 
            $s = s_{t+1}$
        }
        
    }
    \Return{$fitness$}\
}

\end{algorithm}

\vspace{-5pt}
\section{Experiments}
We conducted experiments using three widely used single-domain datasets for DP research: movie-ticket booking, restaurant reservation, and taxi ordering, based on the Microsoft Dialogue Challenge platform \citep{li2017end,li2016user,li2018microsoft,zhao2024decomposed,niu2024scheduled}. To assess the generalizability of EIERL, we also performed experiments on the MultiWOZ2.1 dataset \citep{budzianowski2018multiwoz}, which includes seven domains and is available on the ConvLab platform. 

The objectives of experiments are to validate: 1) the significant advantage of EIERL in balancing exploration-exploitation (\autoref{sec:result1}); 2) the notable improvement of the EII mechanism in accelerating EA search efficiency (\autoref{sec:result2}); 3) the impact of EA-introduced hyperparameters on performance (\autoref{sec:result3}); 4) the independent contributions of RL and EA to EIERL (\autoref{sec:result4}); 5) the generalizability of EIERL on multi-domain tasks\footnote{Our code is publicly available at \nolinkurl{https://github.com/niulinbiao/eierl}.}(\autoref{sec:result5}).

\subsection{Baselines}
We compared our EIERL with publicly available dialogue agent benchmarks, dialogue agents designed to balance exploration and exploitation, and promising LLM-based dialogue agents:

\begin{itemize} 
\item \textbf{DQN$\_$EPSILON$\_N$} agents are trained using standard DQN with a traditional $\epsilon-greedy$ exploration strategy, where $\epsilon=N$ \citep{mnih2015human}\footnote{We evaluated the effect of different $\epsilon$ values on performance across various dialogue domains, selecting the optimal $\epsilon$ value for each domain as the baseline. Detailed results and analysis are provided in \autoref{app:explor}.}.
\item \textbf{NOISY$\_$DQN} agents enhance exploration by introducing noise into the network weights \citep{0005ZLLL22}.
\item \textbf{ICM$\_$DQN} agents incorporate intrinsic curiosity rewards to encourage exploration of the new space \citep{PathakAED17}.
\item LLM-based DA-level (\textbf{LLM$\_$DP}) agents replace the DP module of the TOD system with an LLM, selecting suitable actions to be passed to the NLG for response generation\footnotemark.
\item LLM-based word-level (\textbf{LLM$\_$DP$\_$NLG}) agents replace both the DP and NLG modules of the TOD system with an LLM, directly selecting suitable words to construct responses \citep{abs-2402-18013}\footnotemark[\value{footnote}].
\end{itemize}
\footnotetext{Given the lack of research on replacing DP with LLMs or fine-tuning them for task-oriented DP tasks, and the high computational cost of fine-tuning, we chose GPT-4.0 as LLM agents for its strong generative abilities and popularity. We also evaluated other open-source LLMs such as LLaMA 3 and Gemma2, but their performance was far inferior to ChatGPT-4.0, so only the results of the latter are reported. Tailored prompts are detailed in Appendix~\ref{sec:appendix}.}

\begin{table*}[ht!]
\caption{Evaluation results for all agents across the three benchmark datasets are provided, with the highest value in each metric column highlighted in bold. All results of agent pairs are statistically significant at the same epoch (t-test, p < 0.05). Epochs (50, 250, 500) represent early, mid, and post-convergence training stages. The reason for using 500 epochs as the display cutoff, detailed results and variances are provided in \autoref{app:main}.} 
\centering
\begin{adjustbox}{max width=\textwidth}
\renewcommand{\arraystretch}{0.85} 
\begin{tabular}{cl|ccc|ccc|ccc}
\toprule
\multirow{2}{*}{Domain} &
  \multicolumn{1}{c|}{\multirow{2}{*}{Agent}} &
  \multicolumn{3}{c|}{Epoch = 50} &
  \multicolumn{3}{c|}{Epoch = 250} &
  \multicolumn{3}{c}{Epoch = 500} \\ \cmidrule{3-11} 
 & \multicolumn{1}{c|}{} & \textbf{Success$\uparrow$} & \textbf{Reward$\uparrow$} & \textbf{Turns$\downarrow$} & \textbf{Success$\uparrow$} & \textbf{Reward$\uparrow$} & \textbf{Turns$\downarrow$} & \textbf{Success$\uparrow$} & \textbf{Reward$\uparrow$} & \textbf{Turns$\downarrow$} \\ \midrule
\multirow{7}{*}{Movie}      
& DQN\_EPSILON\_0.0  & 0.3505 & -13.00 & 32.11 & 0.5403 & 12.99  & 25.70 & 0.5553 & 14.95  & 25.37 \\
& DQN\_EPSILON\_0.05 & 0.3093 & -18.61 & 33.44 & 0.6795 & 31.84  & 21.39 & 0.7668 & 43.42  & 19.21 \\
& NOISY\_DQN         & 0.4137 & -4.73  & 30.75 & 0.7141 & 36.68  & 20.04 & 0.7280 & 39.38  & 20.16 \\
& ICM\_DQN           & 0.1475 & -37.81 & 33.00 & 0.5166 & 10.37  & 25.23 & 0.5311 & 12.49  & 24.47 \\
& LLM\_DP            & \textbf{0.4156} & \textbf{-3.09}  & \textbf{27.34} & 0.4156 & -3.09  & 27.34 & 0.4156 & -3.09  & 27.34 \\
& LLM\_DP\_NLG       & 0.2564 & -24.98 & 30.96 & 0.2564 & -24.98 & 30.96 & 0.2564 & -24.98 & 30.96 \\ \cdashline{2-11}[1pt/3pt]
& EIERL              & 0.2372 & -27.53 & 34.01 & \textbf{0.8033} & \textbf{48.21}  & \textbf{18.36} & \textbf{0.8552} & \textbf{55.29}  & \textbf{16.66} \\\midrule

\multirow{7}{*}{Rest.} 
& DQN\_EPSILON\_0.0  & 0.0695 & -36.57 & 27.66 & 0.4907 & 4.10   & 22.13 & 0.5671 & 11.63  & 23.22 \\
& DQN\_EPSILON\_0.05 & 0.0726 & -36.28 & 27.63 & 0.5712 & 12.30  & 20.21 & 0.5817 & 12.79  & 21.12 \\
& NOISY\_DQN         & 0.0000 & -43.92 & 29.84 & 0.1669 & -28.25 & 28.55 & 0.2988 & -15.20 & 26.18 \\
& ICM\_DQN           & 0.0067 & -40.85 & 24.90 & 0.0231 & -38.92 & 23.99 & 0.0082 & -32.88 & \textbf{9.25}  \\
& LLM\_DP            & \textbf{0.3896} & \textbf{-5.96}  & \textbf{20.16} & 0.3896 & -5.96  & 29.16 & 0.3896 & -5.96  & 29.16 \\
& LLM\_DP\_NLG       & 0.2498 & -20.14 & 33.72 & 0.2498 & -20.14 & 33.72 & 0.2498 & -20.14 & 33.72 \\ \cdashline{2-11}[1pt/3pt]
& EIERL              & 0.0181 & -41.09 & 27.44 & \textbf{0.6975} & \textbf{24.79}  & \textbf{17.98} & \textbf{0.7935} & \textbf{34.99}  & 16.07 \\\midrule

\multirow{7}{*}{Taxi}       
& DQN\_EPSILON\_0.0  & 0.0004 & -42.69 & 27.47 & 0.4846 & 2.26   & 24.70 & 0.5879 & 12.38  & 23.06 \\
& DQN\_EPSILON\_0.05 & 0.0000 & -42.86 & 27.71 & 0.5598 & 8.19  & 22.38 & 0.6683 & 20.19  & 21.90 \\
& NOISY\_DQN         & 0.0000 & -43.73 & 29.46 & 0.1455 & -30.56 & 29.32 & 0.2615 & -19.46 & 28.00 \\
& ICM\_DQN           & 0.0008 & -42.34 & 26.84 & 0.0481 & -34.48 & \textbf{19.62} & 0.0706 & -28.59 & \textbf{11.90} \\
& LLM\_DP            & \textbf{0.3496} & \textbf{-10.23} & 25.95 & 0.3496 & -10.23 & 25.95 & 0.3496 & -10.23 & 25.95 \\
& LLM\_DP\_NLG       & 0.2395 & -19.96 & 30.96 & 0.2395 & -19.96 & 30.96 & 0.2395 & -19.96 & 30.96 \\ \cdashline{2-11}[1pt/3pt]
& EIERL              & 0.0000 & -41.55 & \textbf{25.10} & \textbf{0.5638} & \textbf{9.26}   & 21.96 & \textbf{0.8159} & \textbf{35.39}  & 17.29 \\\bottomrule
\end{tabular}
\end{adjustbox}
\label{tab:main}
\end{table*}

\vspace{-8pt}
\subsection{Implementation Details}
\vspace{-3pt}

Each DQN variant employs a multilayer perceptron with two hidden layers of 80 units apiece, and applies ReLU activations throughout to enhance non-linear expressiveness and mitigate vanishing gradients. DQN$\_$EPSILON$\_$N uses the optimal $\epsilon$ value ($0.05$) for each domain. Additionally, to assess the algorithm's intrinsic exploration without external guidance, we included a baseline with $\epsilon = 0$. This design facilitates a clear assessment of the exploration efficiency of the improved algorithms. Across all experiments the hyperparameters are fixed as follows: discount factor $\gamma = 0.99$, mini batch size $16$, learning rate $0.001$, and replay buffer capacity 5000 transitions. In single-domain tasks, the EIERL algorithm uses an EA population $P$ of 3, a DRL population of 1, and a mutation strength $\sigma$ of 0.1. For multi-domain tasks, the EA population is increased to 10 and the DRL population to 5, while keeping the mutation strength constant. A successful dialogue yields a reward of $2L$, whereas an unsuccessful one incurs a penalty of $-L$. Each turn also carries a constant cost of $-1$ to encourage concise exchanges. The dialogue length is limited to $L$ turns, with $L = 30$ in single-domain experiments and $L = 40$ in multi-domain experiments. In addition, to ensure a fair comparison among baseline methods, we consistently set the number of training epochs to 500 for single-domain tasks and 10,000 for multi-domain tasks. While certain methods may demonstrate improved performance with increased training epochs, our focus lies in optimizing the balance between exploration and exploitation to enhance learning efficiency, as reflected by achieving higher success rates with fewer training epochs.

Before training begins, all dialogue agents perform a warm start for 120 epochs to pre-fill the experience replay buffer. During the training phase of each epoch, the agents interact with the environment once, storing the acquired experiences in the replay buffer. In the testing phase, each agent interacts with the environment fifty times, averaging these results to produce a one-time result for the current epoch, while the experiences obtained are not stored. Each agent model is executed five times with random seeds, and the average results are used for analysis and comparison to ensure robustness. Furthermore, since our EIERL involves multiple agent individuals in its population, we randomly sample 1/M of the experiences (where M is the number of individuals) from the acquired experiences for storage in the replay buffer, maintaining consistent training costs.

\begin{figure*}[htbp]
\centering

\begin{minipage}[b]{4.8cm}
    \centering
    \includegraphics[width=5cm]{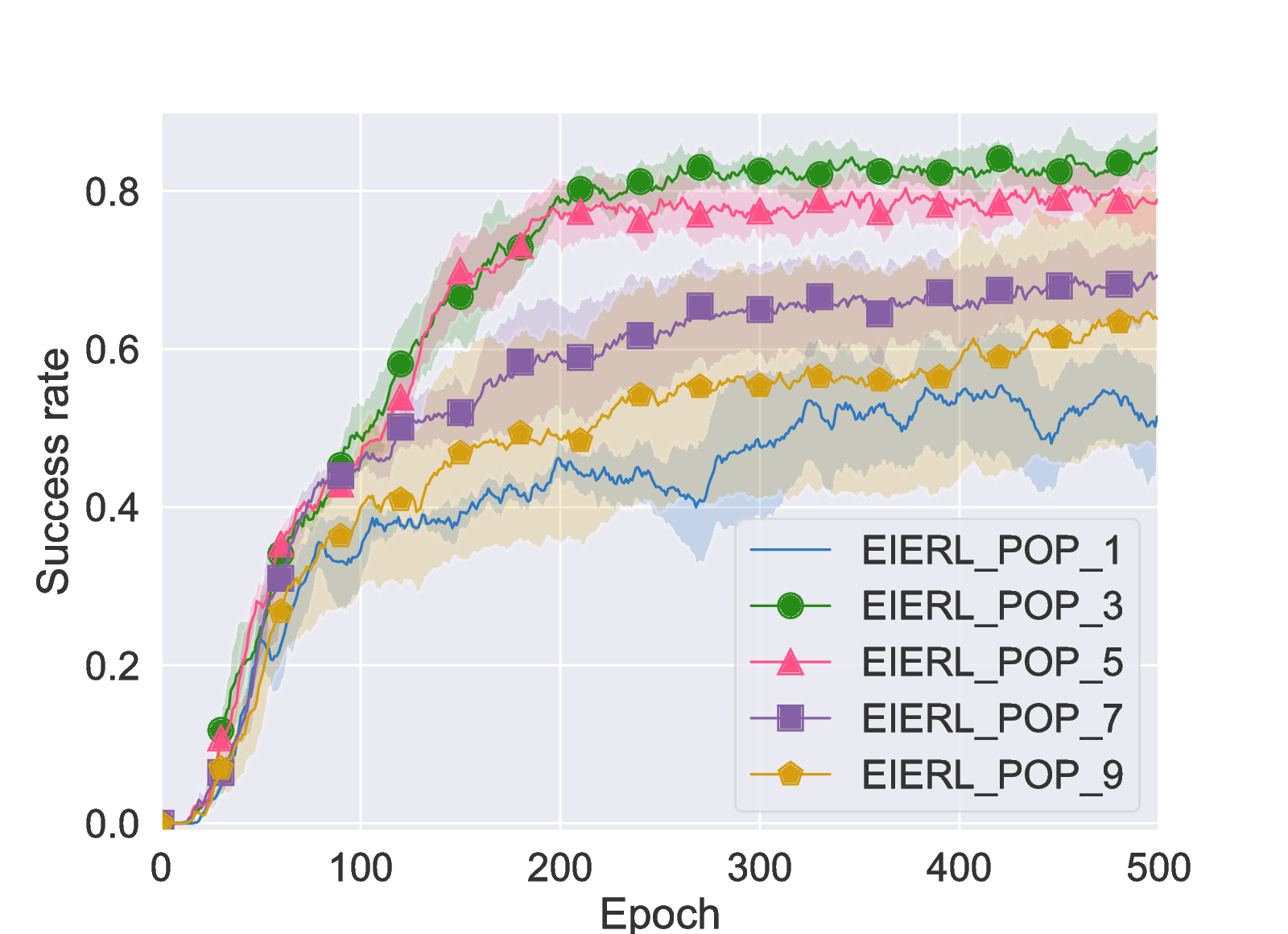}
    \caption*{(a) Movie}
\end{minipage}
\hspace{0.5cm}
\begin{minipage}[b]{4.8cm}
    \centering
    \includegraphics[width=5cm]{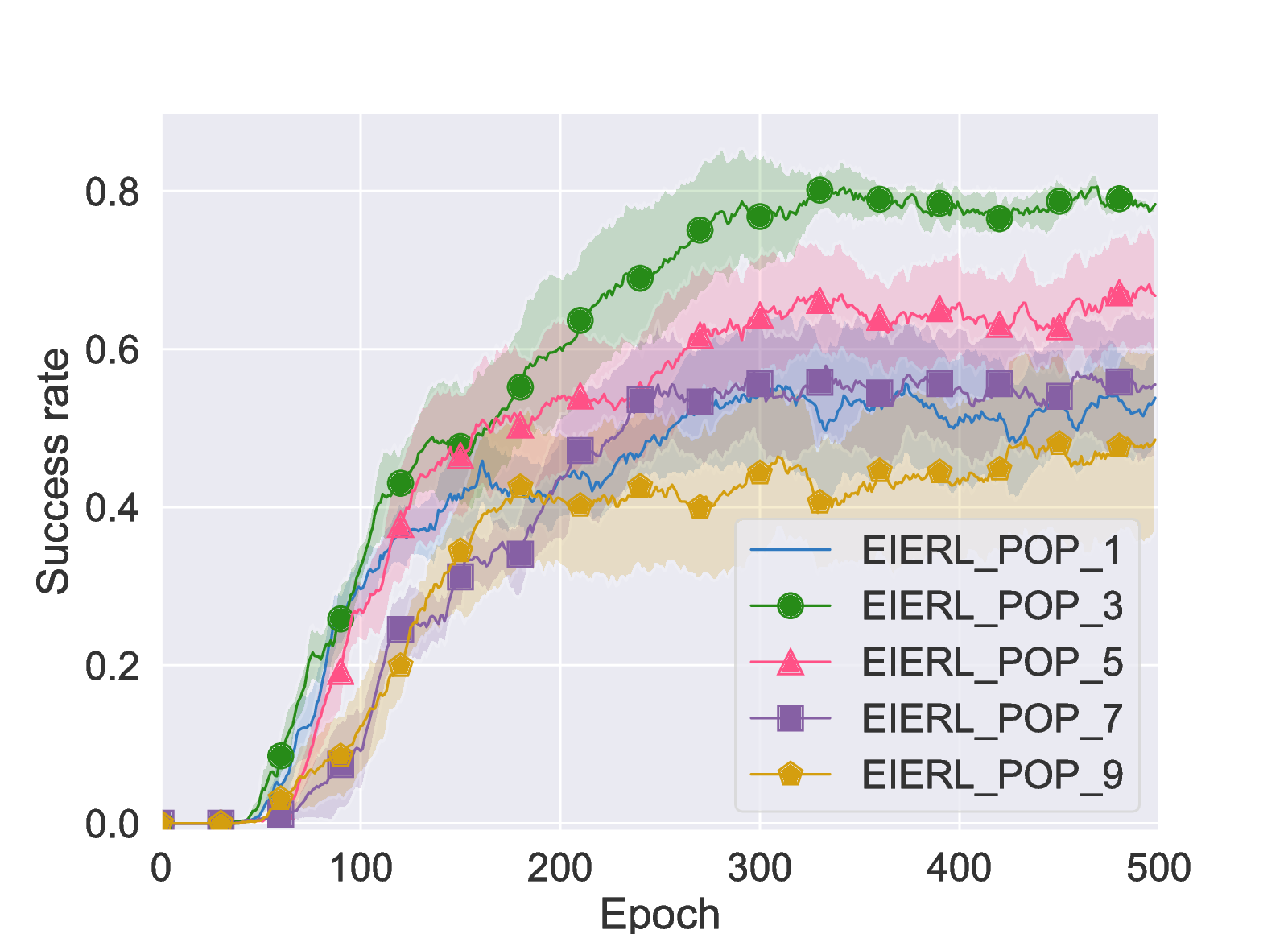}
    \caption*{(b) Rest}
\end{minipage}
\hspace{0.5cm}
\begin{minipage}[b]{4.8cm}
    \centering
    \includegraphics[width=5cm]{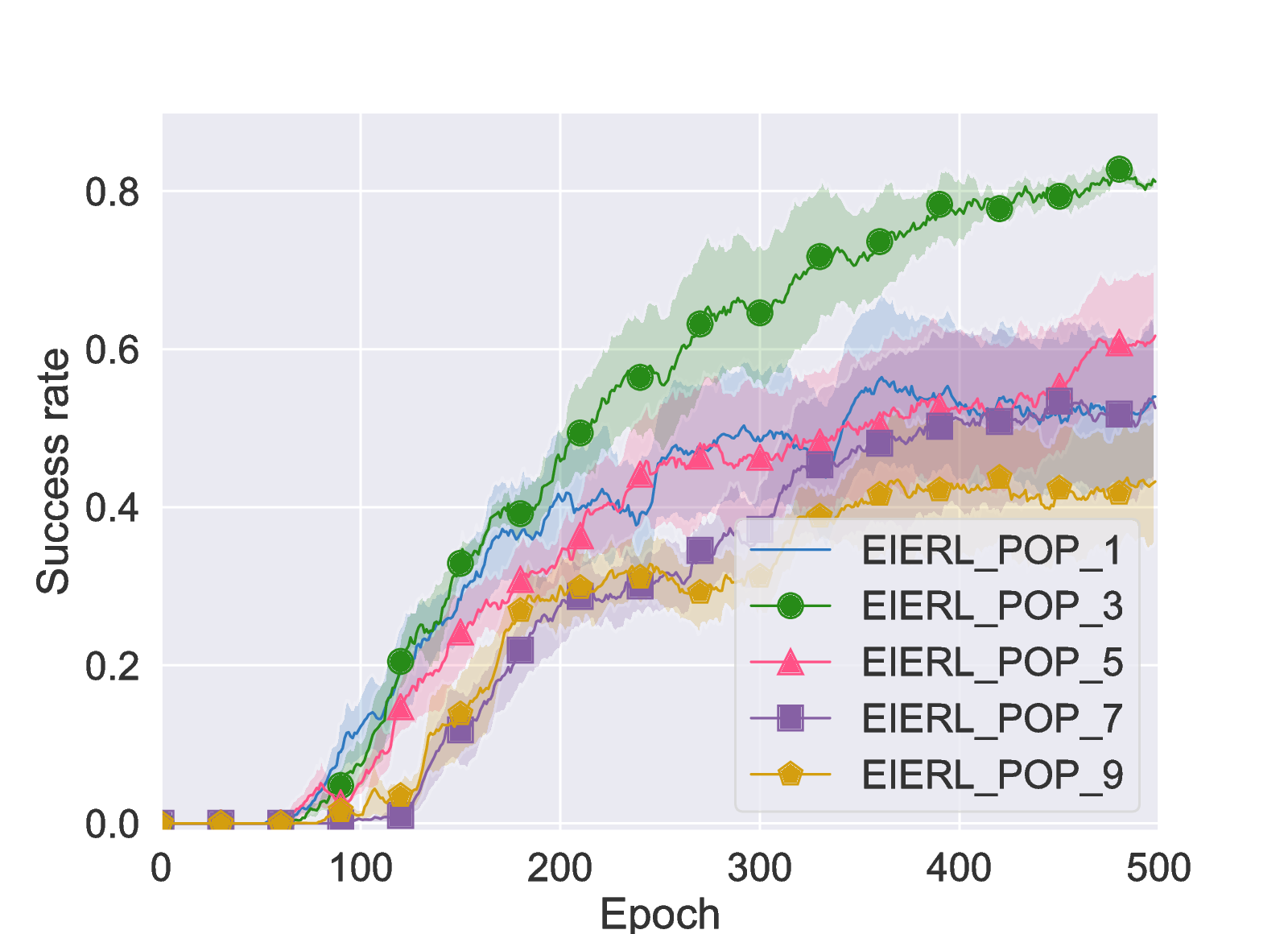}
    \caption*{(c) Taxi}
\end{minipage}

\caption{The effect of the EA population size ($P$) on performance across Movie, Restaurant, and Taxi domains.}
\label{fig:figure5}
\end{figure*}

\begin{figure*}[htbp]
\centering

\begin{minipage}[b]{4.8cm}
    \centering
    \includegraphics[width=5cm]{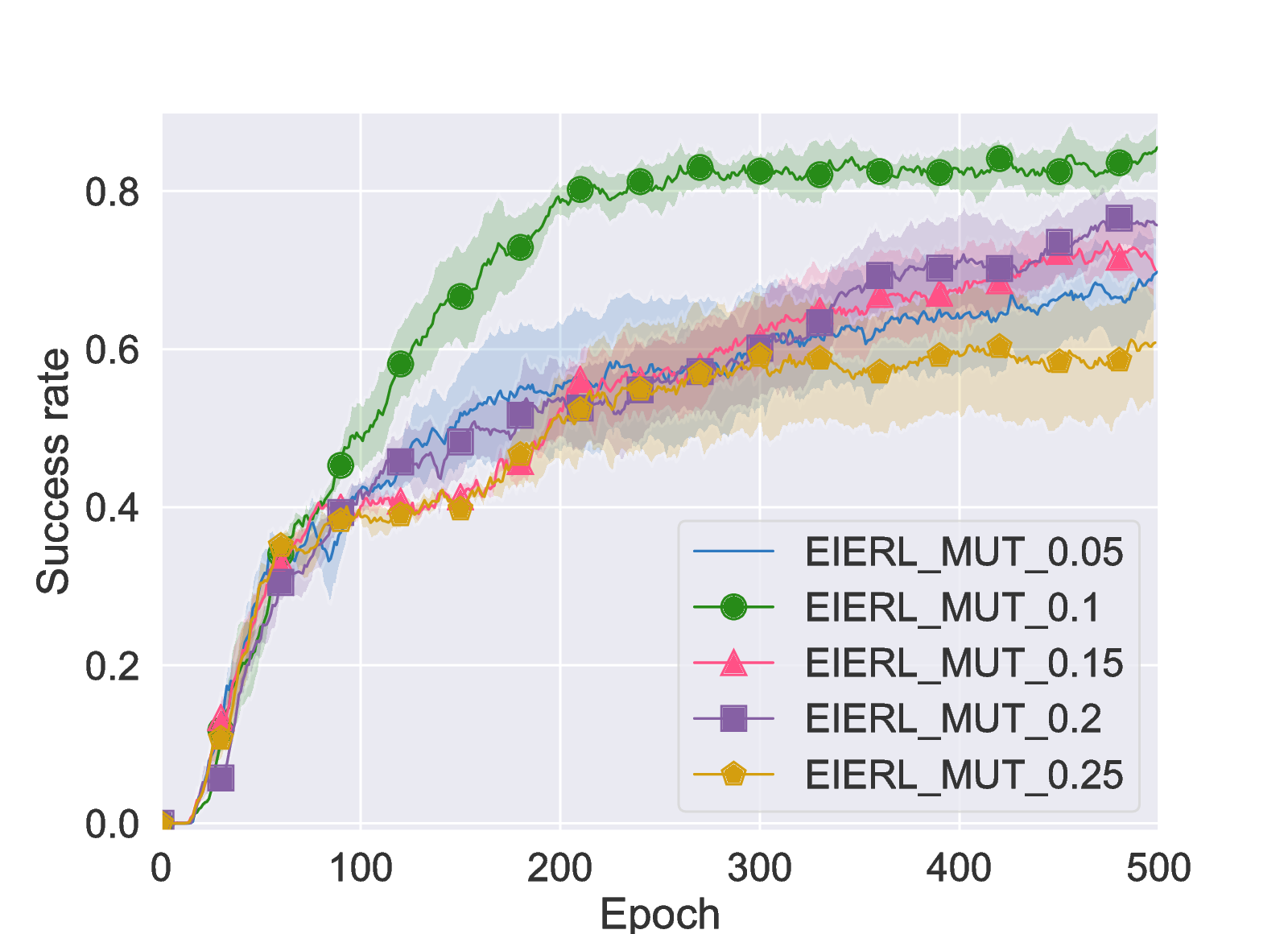}
    \caption*{(a) Movie}
    \label{fig:figure6-1}
\end{minipage}
\hspace{0.5cm}
\begin{minipage}[b]{4.8cm}
    \centering
    \includegraphics[width=5cm]{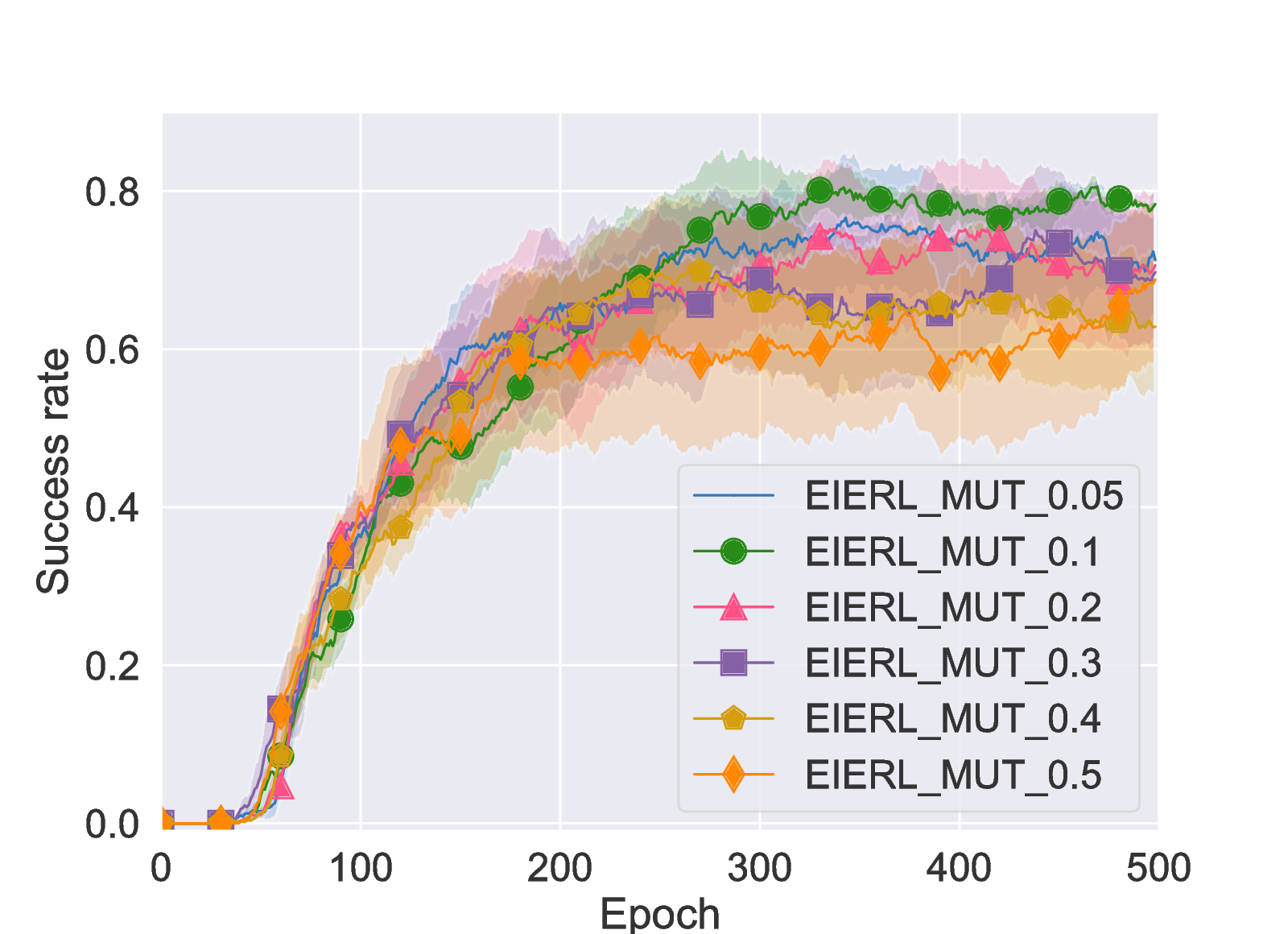}
    \caption*{(b) Rest}
    \label{fig:figure6-2}
\end{minipage}
\hspace{0.5cm}
\begin{minipage}[b]{4.8cm}
    \centering
    \includegraphics[width=5cm]{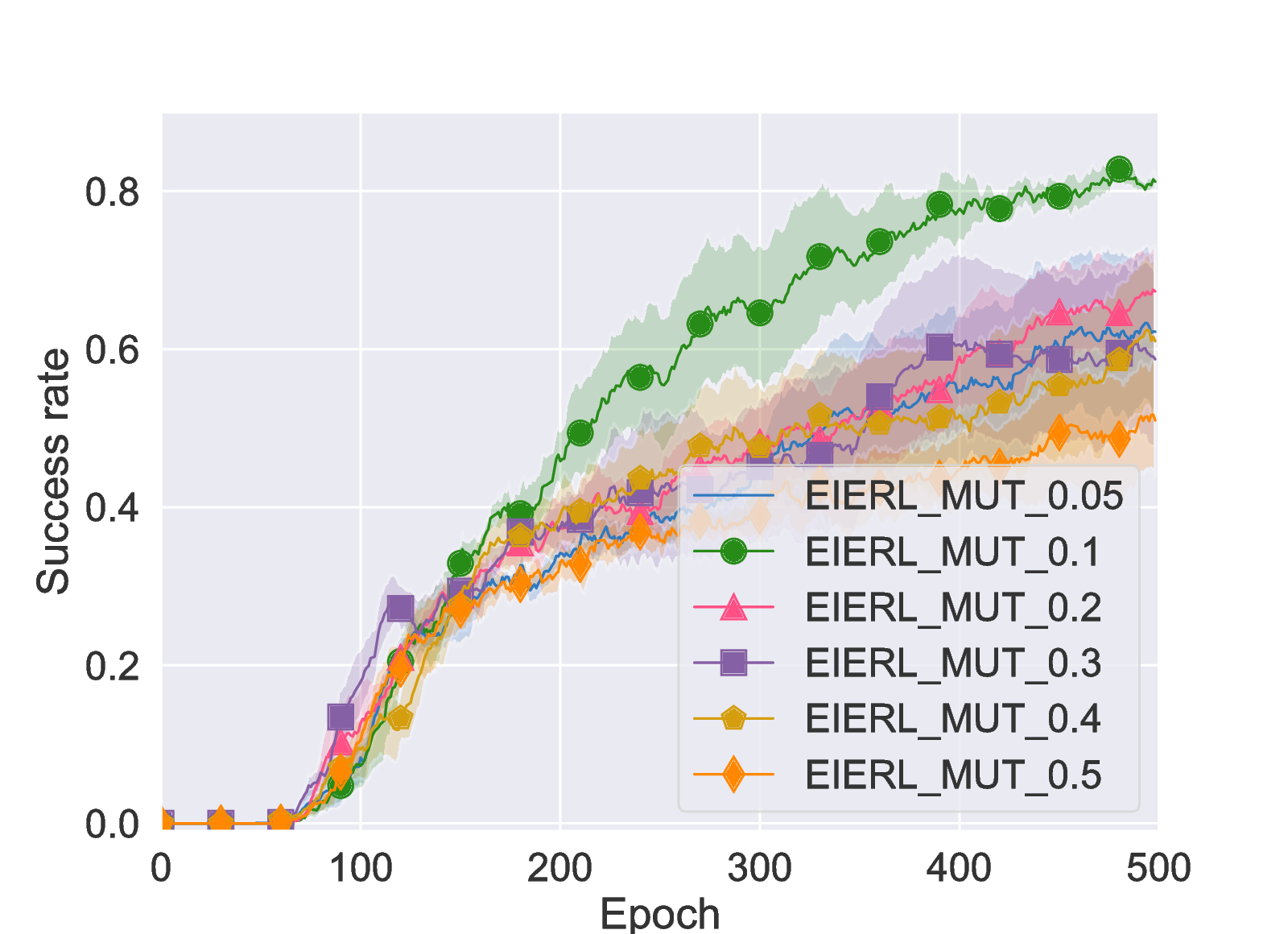}
    \caption*{(c) Taxi}
    \label{fig:figure6-3}
\end{minipage}

\caption{The impact of mutation strength ($\sigma$) on performance across Movie, Restaurant, and Taxi domains.}
\label{fig:figure6}
\end{figure*}

\subsection{Main Results}
\label{sec:result1}
The performance comparison of various baselines across different domains is illustrated in \autoref{tab:main}. The performance contrast between DQN$\_$EPSILON$\_$0.0 and DQN$\_$EPSILON$\_$0.05 highlights that appropriate exploration significantly improves performance. In contrast, although ICM$\_$DQN also provides exploration guidance, its method of encouraging exploration in new state spaces is not well-suited for task-oriented dialogue scenarios with clearly defined goals. While NOISY$\_$DQN performs well in the movie domain, the increased complexity of state spaces in the taxi and restaurant domains, due to the larger number of user goals and slots \citep{ZhaoQWZW22}, exacerbates the problem of exploration-exploitation trade-off, complicating learning process. This challenge is also evident in ICM$\_$DQN, which encourages the exploration of new state spaces. It is worth noting that although ICM\_DQN achieves fewer average dialogue turns in both domains, its success rate and average reward are also lower. This is primarily due to ICM\_DQN encountering many dialogues that fail quickly, leading to fewer turns on average and a lower success rate. Although LM$\_$DP and LLM$\_$DP$\_$NLG with tailored prompts perform well in the early stages by leveraging large-scale LLM data, their lack of fine-tuning for DP tasks prevents further performance improvement. Moreover, due to the closed-source nature of ChatGPT-4.0, epochs cannot be directly used to measure its performance. Consequently, we focus more on comparing the performance of the converged trainable model with that of LLMs. In contrast, our EIERL achieves superior performance across various domains due to a better balance between exploration and exploitation.

\subsection{Analysis}
\subsubsection{Effectiveness of the EII Mechanism}
\label{sec:result2}

To evaluate the effectiveness of the EII mechanism, we compared the performance of EIERL with ERL, which excludes the EII mechanism, across different domains. As shown in Figure 4, EIERL consistently demonstrated a faster convergence rate and greater stability across various domains. In the early training stages, the performances of ERL and EIERL were comparable. However, as training progressed into later stages, EIERL significantly outperformed ERL in both success rate and the smoothness of the learning curve, particularly in the more challenging restaurant and taxi domains. This advantage arises from the EII mechanism, which allows EIERL to effectively harness the high-quality traits of elite individuals, enabling the EA to move toward more optimized search directions and significantly enhancing the exploration process. This mechanism addresses the prolonged evolutionary time challenge associated with incorporating EA into ERL, thereby accelerating the optimization process and minimizing performance fluctuations.

\begin{figure*}[htbp]
\centering

\begin{minipage}[b]{4.8cm}
    \centering
    \includegraphics[width=5cm]{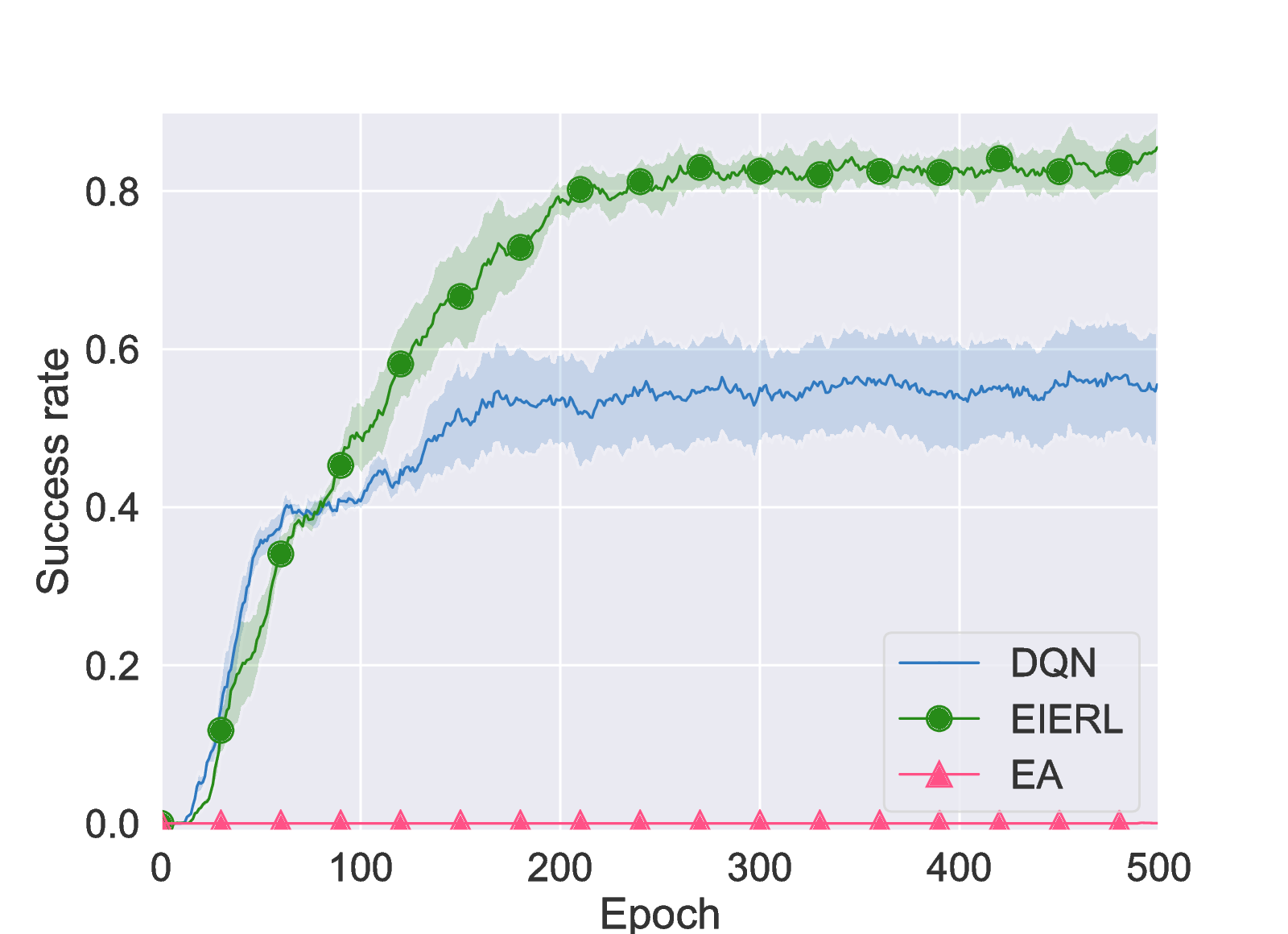}
    \caption*{(a) Movie}
\end{minipage}
\hspace{0.5cm}
\begin{minipage}[b]{4.8cm}
    \centering
    \includegraphics[width=5cm]{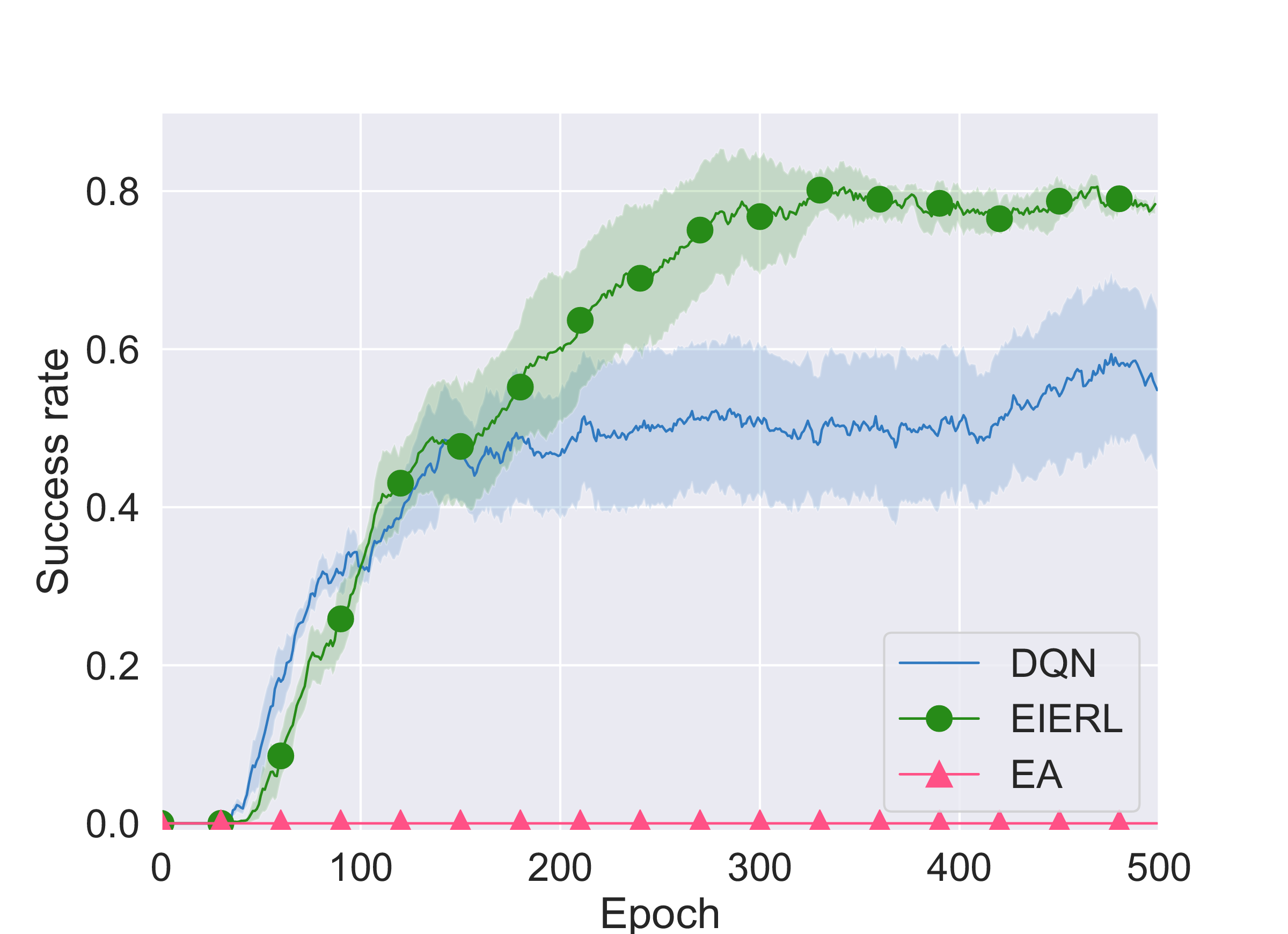}
    \caption*{(b) Rest}
\end{minipage}
\hspace{0.5cm}
\begin{minipage}[b]{4.8cm}
    \centering
    \includegraphics[width=5cm]{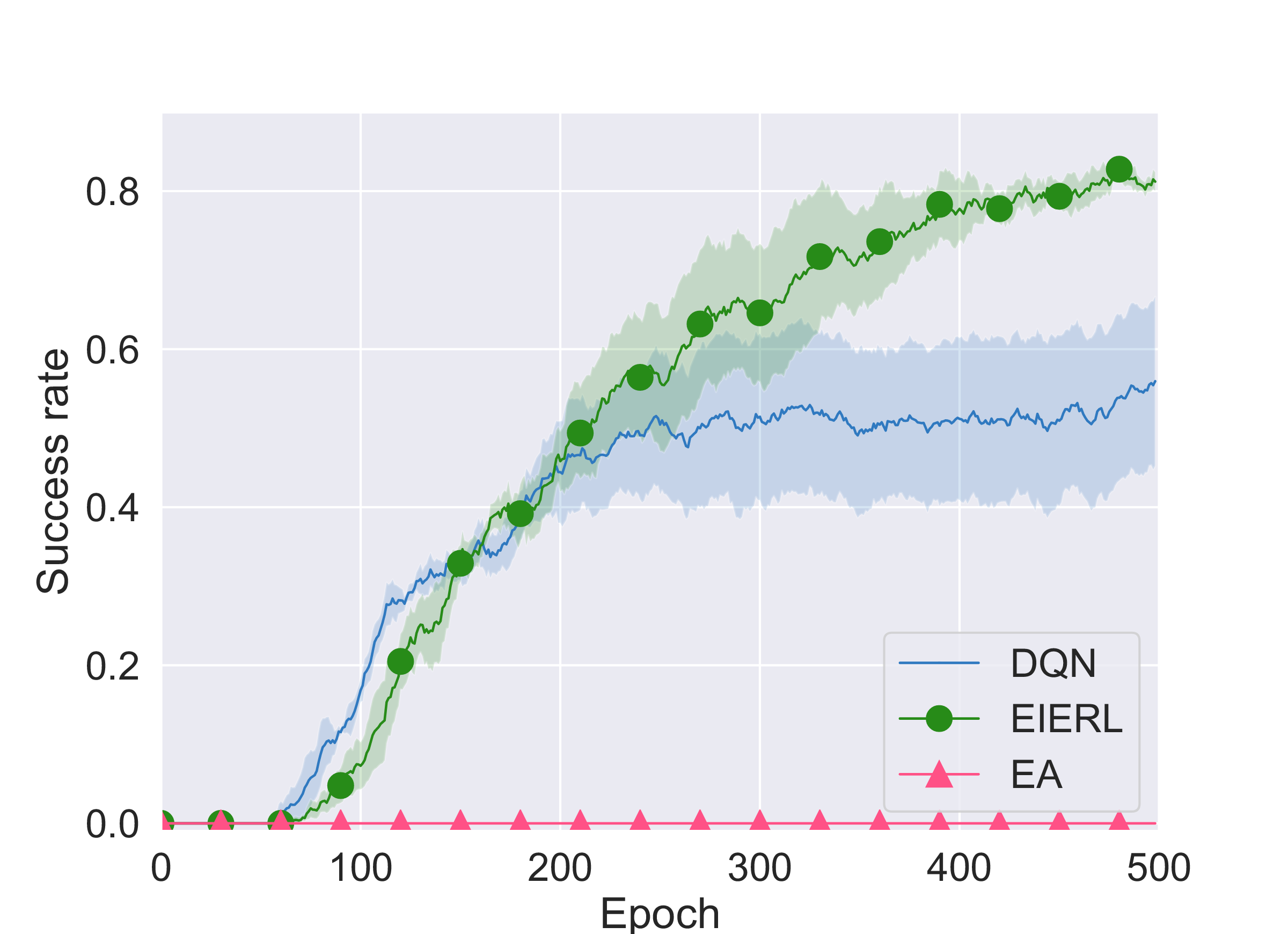}
    \caption*{(c) Taxi}
\end{minipage}

\caption{Ablation experiment of two components (DRL and EA) of our method in Movie, Restaurant, and Taxi domains.}
\label{fig:figure7}
\end{figure*}

\begin{figure*}[htbp]
\centering

\begin{minipage}[b]{4.8cm}
    \centering
    \includegraphics[width=4.8cm]{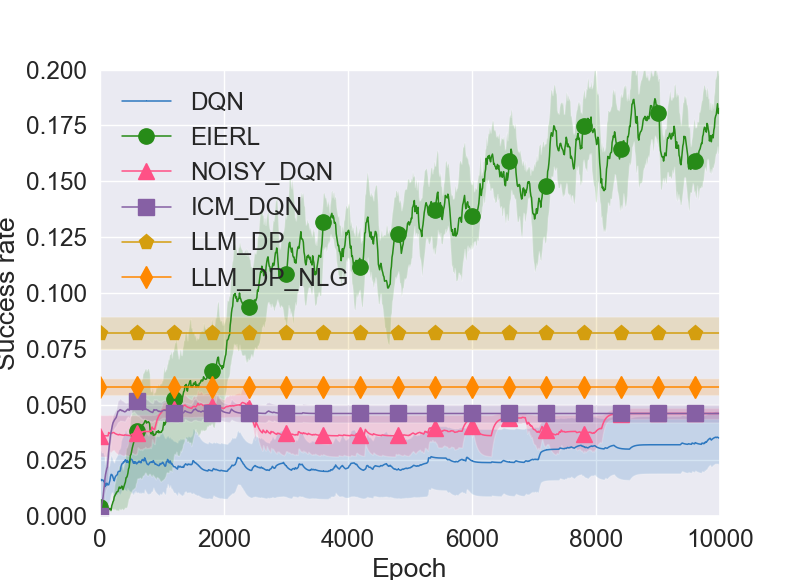}
    \caption*{(a) Effectiveness Experiments.}
\end{minipage}
\hspace{0.5cm}
\begin{minipage}[b]{4.8cm}
    \centering
    \includegraphics[width=4.8cm]{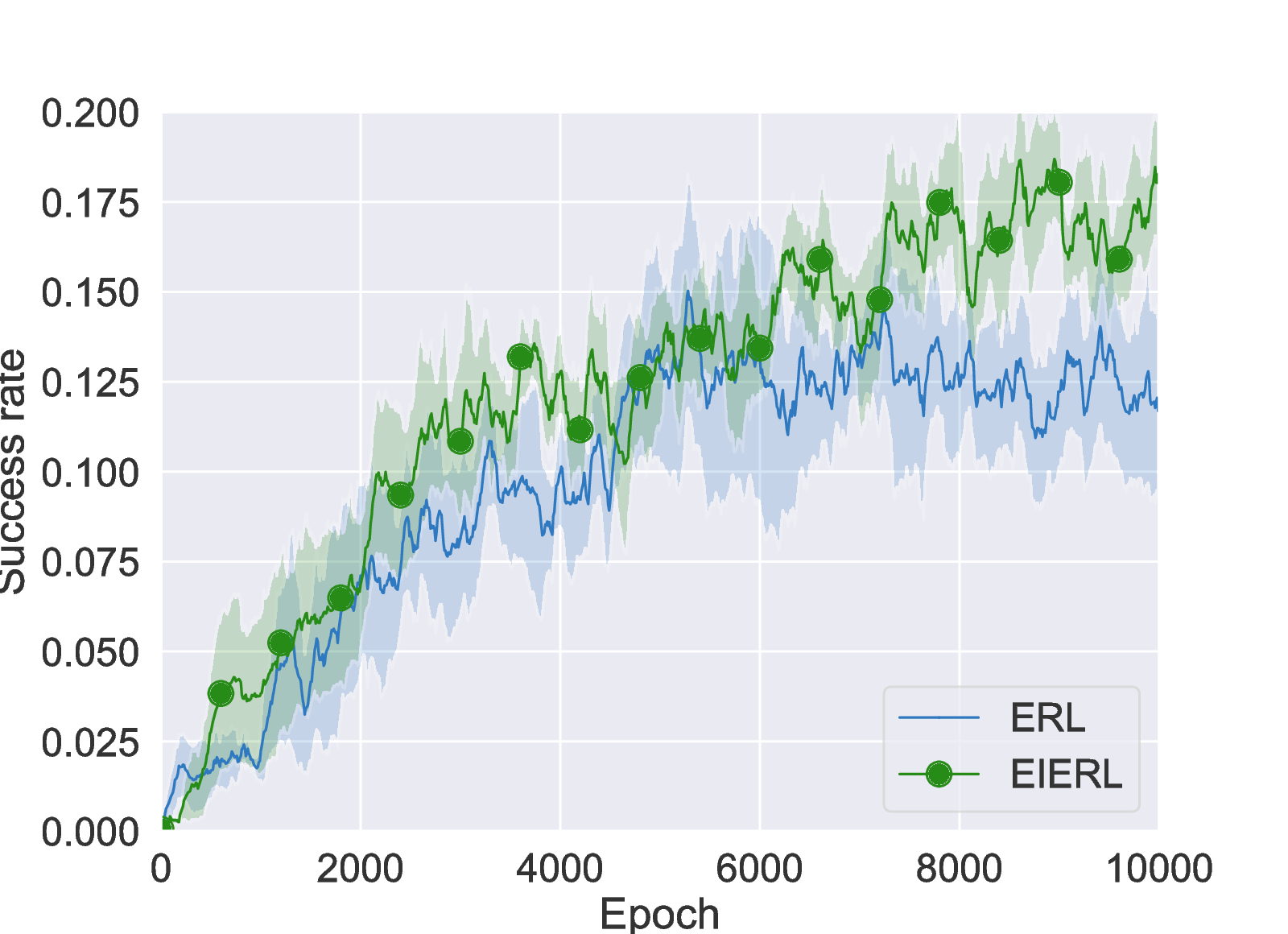}
    \caption*{(b) Effectiveness of EII.}
    \label{fig:figure8(b)}
\end{minipage}
\hspace{0.5cm}
\begin{minipage}[b]{4.8cm}
    \centering
    \includegraphics[width=4.8cm]{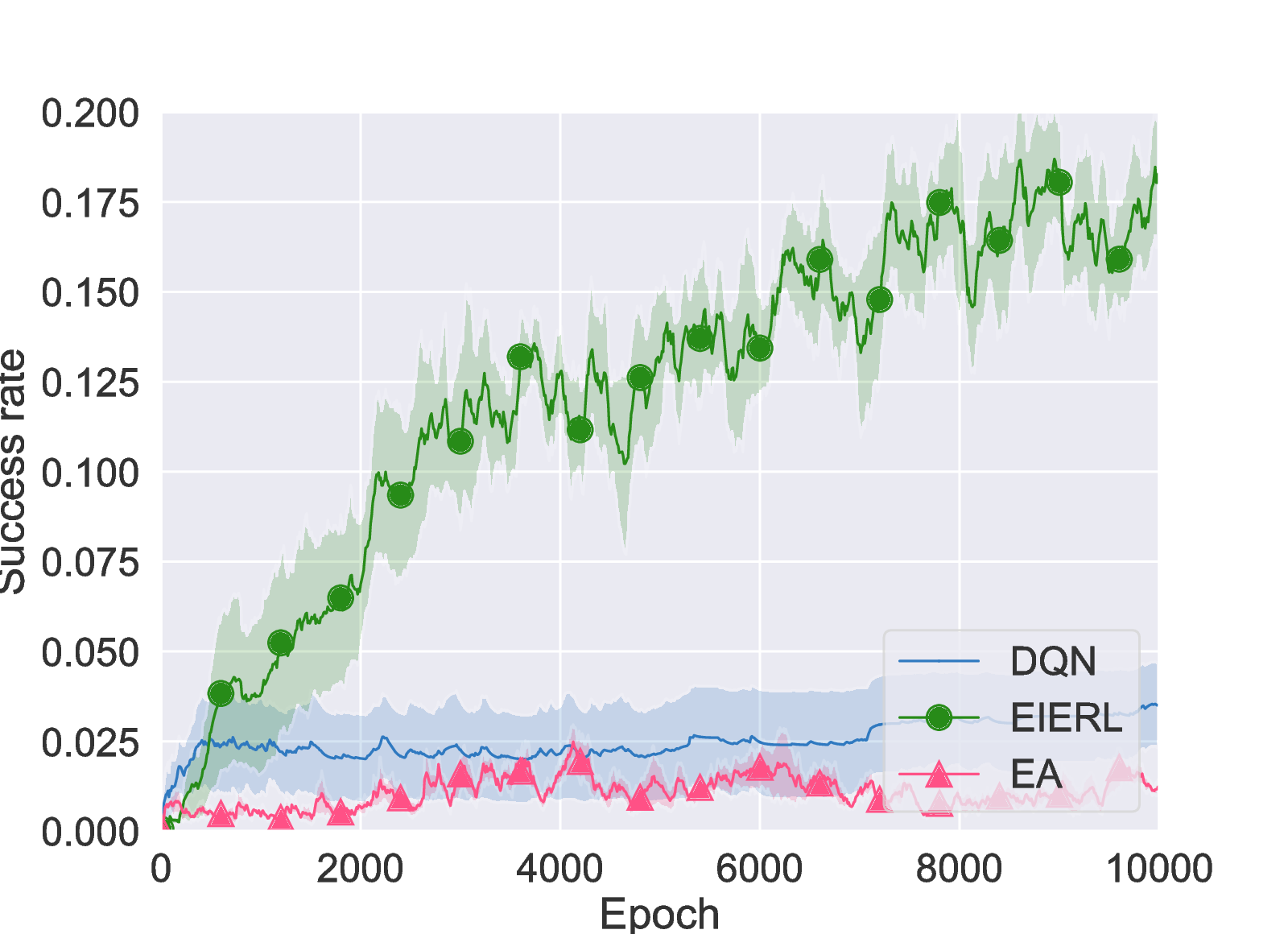}
    \caption*{(c) Ablation Experiment.}
\end{minipage}

\caption{Generalizability experiments of EIERL on MultiWOZ dataset with multiple domains.}
\label{fig:figure8}
\end{figure*}

 \subsubsection{Impact of EA Hyperparameters}
 \label{sec:result3}
The incorporation of EA introduces two key hyperparameters: EA population size ($P$) and mutation strength ($\sigma$). To investigate how these hyperparameters influence EIERL’s performance and provide insights for future practitioners, we conducted two experiments focused on tuning these parameters.

Intuitively, the size of $P$ plays a crucial role in balancing exploration and exploitation. $P$ represents the number of individuals in the EA population. We hypothesized that a larger EA population could lead to excessive exploration, particularly in dialogue tasks with large state spaces, resulting in low-quality experiences. Conversely, a smaller EA population might lead to insufficient exploration, trapping the model in suboptimal policies. To varify this hypothesis, we conducted experiments with different values of $P$ (denoted as EIERL$\_$POP$\_$P). As shown in \autoref{fig:figure5}, both excessively large and small populations negatively impacted performance, with the model performing best when $P$ was set to 3. These findings confirm our hypothesis, and we set $P = 3$ as the default for all experiments, except in this experiment.

Similarly, mutation strength ($\sigma$) affects the variability and quality of individuals in the EA population, which directly influences the quality of the exploration experience. We hypothesized that a large $\sigma$ would result in overly random mutations, decreasing exploration quality, while a small $\sigma$ would limit the EA’s ability to explore the solution space effectively. To test this, we experimented with different values of $\sigma$ (denoted as EIERL$\_$MUT$\_\sigma$)\footnote{Given the larger state spaces in the restaurant and taxi domains, a larger step size of 0.1 was used for $\sigma$, while in the smaller movie domain, a step size of 0.05 was applied.}. The results, presented in \autoref{fig:figure6}, validated our hypothesis: a moderate value of $\sigma = 0.1$ consistently yielded the best performance, while both larger and smaller values of $\sigma$ led to suboptimal outcomes. As a result, $\sigma = 0.1$ was selected as the default for all experiments, except in this experiment.

\vspace{-5pt}
\subsubsection{Ablation Experiment}
\label{sec:result4}
To validate the individual contributions of DRL and EA within EIERL, we conducted an ablation study. The results, shown in \autoref{fig:figure7}, demonstrate that EIERL consistently outperforms either DQN or EA alone. DQN exhibited slow learning and ultimately converged to suboptimal policies, highlighting its limitations in exploration efficiency. In contrast, EA showed minimal improvement throughout the training process. We attribute this to the challenges traditional EA faces in dialogue tasks with high-dimensional state spaces, which require a large number of samples for effective exploration. This further emphasizes the significant improvement that the EII mechanism brings to EA. In summary, EIERL excels in complex dialogue tasks by effectively combining the global search capabilities of EA with the local optimization strengths of DQN.

\subsubsection{Generalizability Experiments}
\label{sec:result5}
To validate the generalizability of EIERL, we conducted further experiments on the MultiWOZ dataset, which includes seven domains: restaurant, taxi, hotel, train, attraction, hospital, and police. These experiments included effectiveness assessments, evaluations of the EII mechanism, and an ablation study. The detailed results are presented in \autoref{fig:figure8}. The findings from these three experiments are consistent with those from the single-domain experiments, demonstrating that EIERL maintains optimal performance in multi-domain dialogue tasks and highlighting the critical role of the proposed EII mechanism.

\vspace{-5pt}
\section{Conclusion}
In this paper, we proposed a novel method, EIERL, that harnesses the global search capabilities of EA alongside the local optimization strengths of DRL to effectively balance the exploration-exploitation in dialogue policy learning. To alleviate the prolonged evolutionary times caused by large spaces in dialogue tasks when integrating EA, we introduced an innovative EII mechanism. This mechanism adaptively determines the optimal timing for injecting the fittest individuals into the EA population based on training progress, thereby providing a clear search direction for exploration and enhancing the evolutionary process. Experiments across multiple dialogue datasets demonstrated the superior performance of EIERL and EII in improving EA's exploration efficiency. To the best of our knowledge, this study is the first to effectively integrate EA with DRL specifically for task-oriented dialogue policy tasks.

\section*{Limitation}
Despite the innovative integration of EA and DRL in this work, which demonstrates significant performance advantages in both single-domain and multi-domain dialogue datasets, we limited the assessment of individual fitness within the population to a single criterion: the cumulative reward commonly used in task-oriented dialogue settings. This reliance on a single fitness measure imposes limitations in complex environments that often involve multiple objectives and multidimensional evaluation metrics, as it may not accurately reflect the true performance of individuals. Consequently, we plan to investigate a multi-criteria fitness evaluation method in future work to enhance the adaptability of our approach in more complex domains. Additionally, future research could combine EAs with advanced reinforcement learning algorithms such as DDPG \citep{lillicrap2015continuous} and SAC \citep{haarnoja2018soft}, which have shown exceptional performance in video game tasks. This combination may lead to further improvements in dialogue tasks.

\section*{Acknowledgments}
We thank the anonymous reviewers for their constructive feedback. Special thanks to the human evaluators who assisted in the manual evaluation of our models. This work was supported by the Hunan Provincial Natural Science Foundation (Grant No. 2024JJ6062), and the National Natural Science Foundation of China (NSFC) via grant 62306342. This work was sponsored by the Excellent Young Scientists Fund in Hunan Province (Grant  No.2024JJ4070), the Science and Technology Innovation Program of Hunan Province under Grant 2024RC3024.

\bibliography{mycustom}
\bibliographystyle{acl_natbib}

\appendix
\newpage  
\onecolumn

\section{Prompt Design for LLM-based Baselines}
\label{sec:appendix}
\vspace{-0.4cm}


\begin{table*}[h]
 \centering
 \caption{Descriptions of Prompts used for LLM-based baselines}
\begin{adjustbox}{width=\linewidth}  
\begin{tabularx}{\textwidth}{|>{\centering\arraybackslash}m{3cm}|>{\centering\arraybackslash}X|}

\hline

\textbf{Model} & \textbf{Prompt} \\
\hline

\textbf{LLM\_DP} & 
\begin{tabular}[t]{@{}m{11.5cm}@{}} 
You must strictly execute the following commands:\\
1. Command execution requirements: when receiving a command, you must strictly follow the given instructions without performing any actions outside the scope of the command or generating any additional words.\\
2. Datasets and system roles: as the dialogue policy component in a task-oriented dialog system, you will make system decisions based on the MultiWOZ 2.1 dataset.\\
3. Processing user dialog state: you will receive a formatted user dialog state: \autoref{lst:json1}. This state will be used as a basis for decision making.\\
4. Generate system actions: based on the user dialog state, you need to generate system actions. These actions should be provided in the following format: [[“ActionType”, “Domain”, “Slot”, “Value”]] where `ActionType` denotes the type of action (e.g. Request, Inform, Confirm, etc.), `Domain` specifies the associated domain (e.g. restaurant, taxi, hotel, etc.), `Slot` is the specific information slot associated with the action (e.g. name, area, type, etc.), and `Value` is the corresponding value or an empty string.
\end{tabular} \\
\hline
\textbf{LLM\_DP\_NLG}  \centering & 
\begin{tabular}[t]{@{}m{11.5cm}@{}} 
You must strictly execute the following commands:\\
1. Command execution requirements: when receiving a command, you must strictly follow the given instructions without performing any actions outside the scope of the command or generating any additional words.\\
2. Datasets and system roles: as the dialogue policy component and the natural language generation component of a task-based dialog system, you will make system decisions based on the MultiWOZ 2.1 dataset.\\
3. Processing user dialog state: you will receive a formatted user dialog state: \autoref{lst:json1}. This state will be used as a basis for decision making.\\
4. Generate system actions: generate your response to user conversations directly.
\end{tabular} \\
\hline
\end{tabularx}
\end{adjustbox}
\end{table*}

\vspace{-0.5cm}
\begin{lstlisting}[language=json,caption={Data Format 1}, label=lst:json1][H]
{
  'user_action': [["Inform", "Hotel", "Area", "east"], ["Inform", "Hotel", "Stars", "4"]],
  'system_action': [],
  'belief_state': {
    'police': {'book': {'booked': []}, 'semi': {}},
    'hotel': {'book': {'booked': [], 'people': '', 'day': '', 'stay': ''},
              'semi': {'name': '', 'area': 'east', 'parking': '', 'pricerange': '', 'stars': '4', 'internet': '', 'type': ''}},
    'attraction': {'book': {'booked': []}, 'semi': {'type': '', 'name': '', 'area': ''}},
    'restaurant': {'book': {'booked': [], 'people': '', 'day': '', 'time': ''},
                   'semi': {'food': '', 'pricerange': '', 'name': '', 'area': ''}},
    'hospital': {'book': {'booked': []}, 'semi': {'department': ''}},
    'taxi': {'book': {'booked': []},
             'semi': {'leaveAt': '', 'destination': '', 'departure': '', 'arriveBy': ''}},
    'train': {'book': {'booked': [], 'people': ''},
              'semi': {'leaveAt': '', 'destination': '', 'day': '', 'arriveBy': '', 'departure': ''}}
  },
  'request_state': {},
  'terminated': False,
  'history': []
}

\end{lstlisting}

\section{Detailed Description of the EIERL Algorithm}
\label{alg:eierl}

\begin{algorithm}[H]

\DontPrintSemicolon
\caption{Elite-Infused Evolutionary Reinforcement Learning}
\label{alg:eierl2}
\KwIn{$N,n,m$}
\KwOut{$Q_{\theta_Q}(s, a)$}

initialize $Q_{\theta_Q}(s, a)$, $Q'_{\theta_{Q'}}(s, a)$ with $\theta_{Q'} = \theta_Q$\;
initialize populations $pop_{policy}$ ($n$) and $pop_{evo}$ ($m$)\;
initialize experience buffer $\mathcal{D}$ and $f_{max} = -\infty$\;

\For{$epoch=1$ \KwTo $N$}{
    initialize $f'_{max} = -\infty$, $\pi_{max}$, and \texttt{$all\_fitness$}\;

  \For{each $\pi \in pop_{policy}$}{load $\theta_Q$ into $\pi$\;}

    \For{each $\pi \in pop_{evo} \cup pop_{policy}$}{
    fitness, $\mathcal{D}$ = Evaluate($\pi$, $\mathcal{D}$)\;
    all\_fitness.append(fitness)\;
    \If{fitness > $f'_{max}$}{$f'_{max} = $ fitness, $\pi_{max} = \pi$\;}
}

   \eIf{$f'_{max} > f_{max}$}{
    replace $pop_{evo}$ with $\pi_{max}$\;
    $f_{max} = f'_{max}$\;
}{
    $pop_{evo}$ = Evolution($pop_{evo}$, $all\_fitness$)\;
}

    sample minibatches of $(s_i, a_i, r_i, s_{i+1})$ from $\mathcal{D}$\
    
    by \autoref{gongshi2} compute $y_i$\
    
    update $\theta_{Q}$ by \autoref{gongshi1}
    
    update the target network $\theta_{Q'} \gets \theta_{Q}$ 
   }

\end{algorithm}

The detailed description of EIERL is shown in \autoref{alg:eierl2}. The EIERL algorithm requires setting a specific number of training epochs, denoted as $N$, which ensures that all baselines converge on the dataset for comprehensive comparison. Additionally, the sizes of the DRL and EA populations, $n$ and $m$ respectively, also should be defined. The goal of EIERL is to obtain a trained DRL agent $Q_{\theta_Q}(s, a)$ that effectively balances exploration and exploitation.

Before training begins, the two Q-values in the DRL agent's network, $Q_{\theta_Q}(s, a)$ and $Q'_{\theta_{Q'}}(s, a)$, are initialized. The experience replay buffer $D$ is pre-filled using experiences obtained through a warm start, and the elite discriminator's initial elite threshold $f_{max}$​ is set to negative infinity. In each epoch, every individual in the DRL population, denoted as ${pop}_{policy}$, duplicates the weight parameters of the trained DRL agent. Both the DRL population ${pop}_{policy}$ and the EA population ${pop}_{evo}$ are evaluated by the elite discriminator within the EII mechanism to assess their fitness. The current maximum fitness, $f'_{max}$, and its corresponding individual $\pi$​ are recorded as $\pi_{max}$. If $f'_{max}$ exceeds the elite threshold $⁡f_{max}$​, the individual is designated as an elite individual, and its fitness​ is updated to $f'_{max}$. The elite injection process is then triggered, where the elite individual $\pi_{max}$​'s weight parameters are loaded into every individual in the EA population ${pop}_{evo}$. If no elite individual is identified, only the EA process (selection, crossover, and mutation) is executed to form the EA population. This process repeats until the specified number of epochs N is reached. Throughout, both EA and DRL populations interact with the environment, generating diverse experiences used for training the DRL agent and evaluating fitness.


\section{Impact of Exploration Degree on Dialogue Policy Learning}
\label{app:explor}
\begin{figure*}[htbp]
\centering

\begin{minipage}[t]{4cm}
    \centering
    \includegraphics[width=4cm]{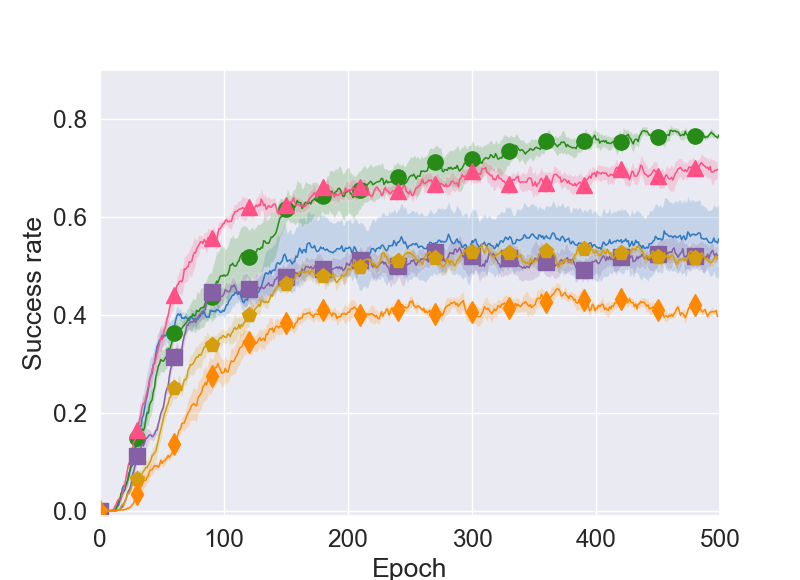}
    \caption*{(a) Movie}
\end{minipage}
\hspace{0.05cm}
\begin{minipage}[t]{4cm}
    \centering
    \includegraphics[width=4cm]{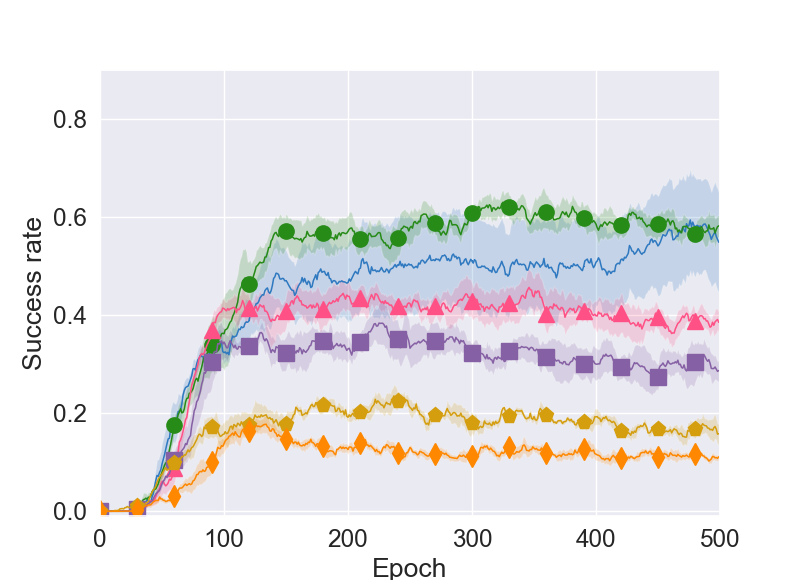}
    \caption*{(b) Rest}
\end{minipage}
\hspace{0.05cm}
\begin{minipage}[t]{4cm}
    \centering
    \includegraphics[width=4cm]{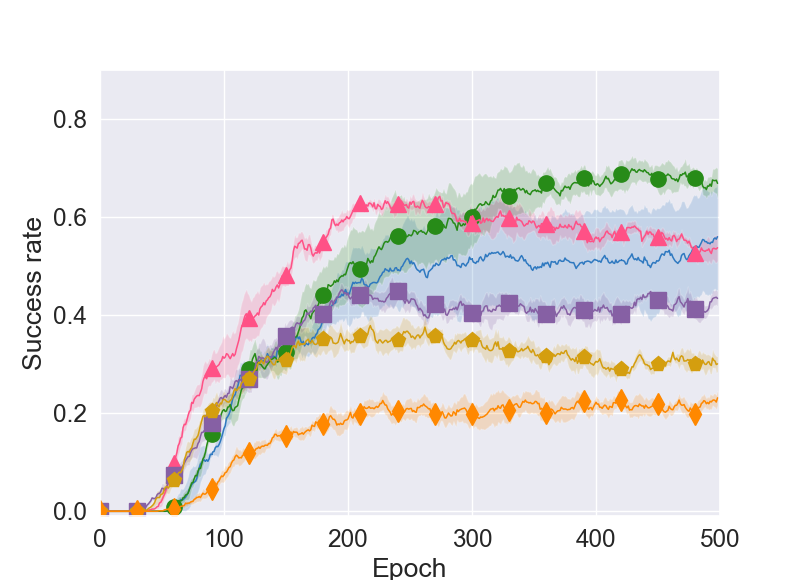}
    \caption*{(c) Taxi}
\end{minipage}
\hspace{0.05cm}
\begin{minipage}[t]{2.8cm}
    \centering
    \includegraphics[width=2.8cm]{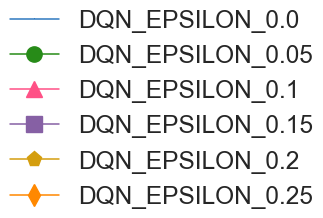}
\end{minipage}

\caption{Effect of epsilon parameters on DQN performance}
\label{fig:figure9}
\end{figure*}

To validate the impact of exploration degree on DRL-based dialogue agents and identify the optimal $\epsilon$ value, we conducted experiments based on the DQN framework, with results presented in \autoref{fig:figure9}. As shown, when exploration is absent, DQN$\_$epsilon$\_0.0$ consistently selects the action with the highest known reward, preventing it from discovering the global optimal policy and causing it to become stuck in a suboptimal strategy.  In contrast, enabling exploration allows the agent to try different actions, thereby uncovering higher-reward pathways. Notably, DQN$\_$epsilon$\_0.05$ achieved the best performance, while performance declined as $\epsilon$ increased beyond this threshold. This highlights that excessive exploration can lead to random action selection, which in turn reduces the quality of experiences. In summary, DRL-based dialogue agents struggle to balance exploration and exploitation, as both excessive and insufficient exploration can negatively affect their performance. Consequently, the best-performing DQN$\_$epsilon$\_0.05$ is used as our baseline model.

\section{Complete Results and Variances with 500 Epochs Display Cutoff}
\label{app:main}
The reason for choosing 500 epochs is to have a better view of the pre-learning curve, and at this point all methods have fully converged. Moreover, our approach aims to optimize the balance between exploration and exploitation, and the primary outcome is improved learning efficiency, which is evidenced by higher success rates achieved within fewer epochs. For consistency and clarity in comparison, all plots and statistical analyses were standardized to 500 epochs.

\begin{figure*}[htbp]
\centering

\begin{minipage}[t]{4cm}
    \centering
    \includegraphics[width=4cm]{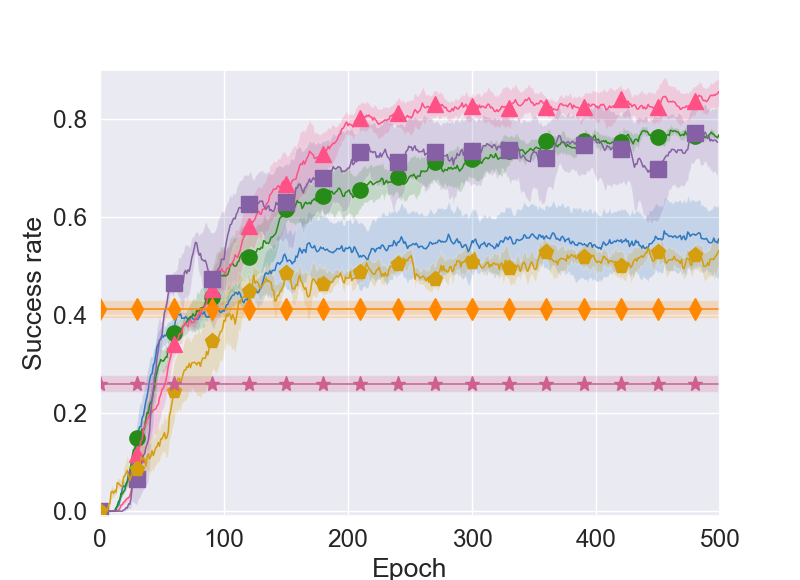}
    \caption*{(a) Movie}
\end{minipage}
\hspace{0.05cm}
\begin{minipage}[t]{4cm}
    \centering
    \includegraphics[width=4cm]{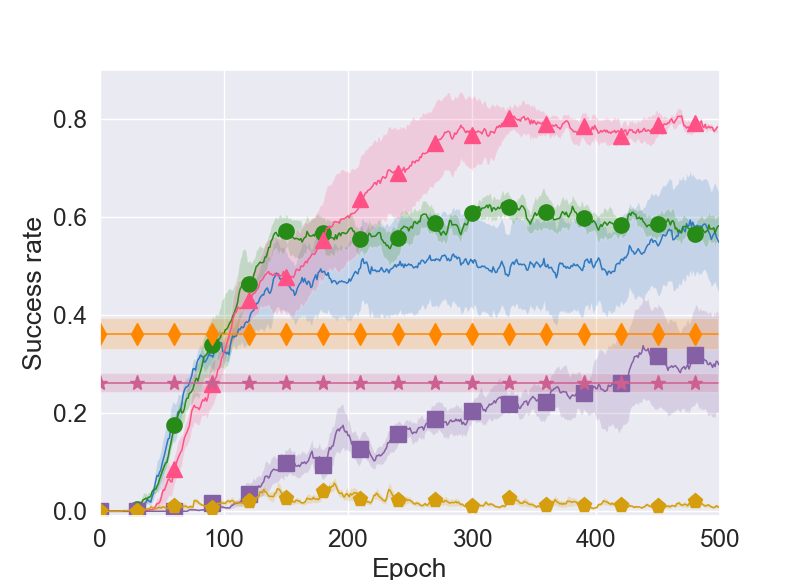}
    \caption*{(b) Restaurant}
\end{minipage}
\hspace{0.05cm}
\begin{minipage}[t]{4cm}
    \centering
    \includegraphics[width=4cm]{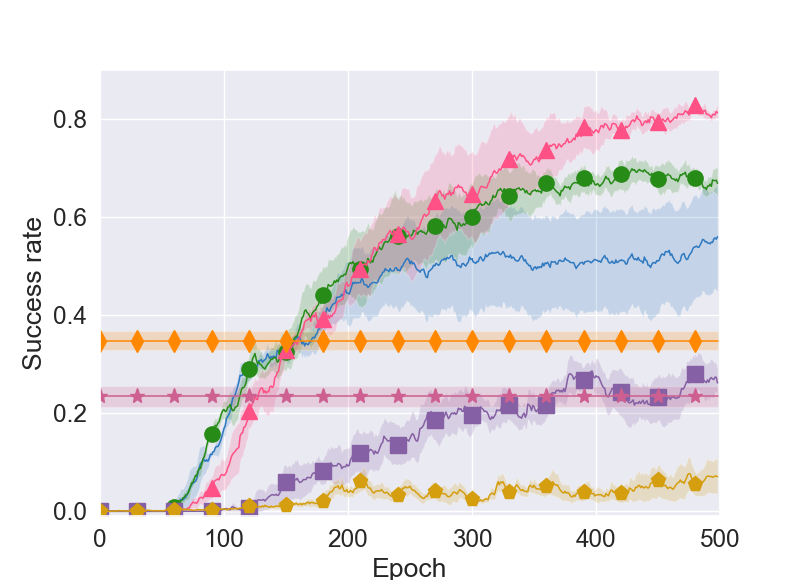}
    \caption*{(c) Taxi}
\end{minipage}
\hspace{0.05cm}
\begin{minipage}[t]{2.8cm}
    \centering
    \includegraphics[width=2.8cm]{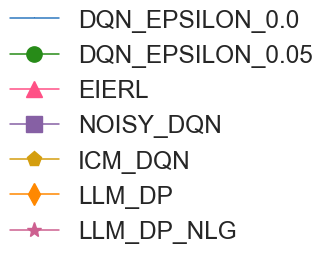}
\end{minipage}

\caption{The learning curves of different agents in Movie, Restaurant, and Taxi domains.}
\label{fig:figure3}
\end{figure*}

\end{CJK}
\end{document}